\documentclass[10pt,twocolumn,letterpaper]{article}

\usepackage{cvpr}
\usepackage{times}
\usepackage{epsfig}
\usepackage{graphicx}
\usepackage{amsmath}
\usepackage{amssymb}

 \usepackage[caption=false,font=footnotesize]{subfig}
\usepackage{booktabs}
\usepackage[british,american]{babel}
\usepackage{amsfonts}
\usepackage{xcolor}
\usepackage{enumitem}
\usepackage[pagebackref=false,breaklinks=true,letterpaper=true,colorlinks,bookmarks=false]{hyperref}
\usepackage{booktabs}
\usepackage{amsfonts}
\usepackage{xcolor}
\usepackage{textpos}
\newcommand{\parbf}[1]{\vspace{0.2cm}\noindent\textbf{#1}}
\newcommand{\figref}[1]{\mbox{Fig.~\ref{#1}}}
\newcommand{\tblref}[1]{\mbox{Table~\ref{#1}}}
\newcommand{\secref}[1]{\mbox{Sec.~\ref{#1}}}
\renewcommand{\eqref}[1]{\mbox{Eq.~\ref{#1}}}

\newcommand*{\email}[1]{\tt\small{#1}}
\newcommand*{\affmark}[1][*]{\textsuperscript{#1}}

\cvprfinalcopy

\begin{document}

\title{Modeling Fashion Influence from Photos}

\author{
\setlength{\tabcolsep}{10pt}
\begin{tabular}{cc}
Ziad Al-Halah\affmark[1] & Kristen Grauman\affmark[1,2]\\
\email{ziadlhlh@gmail.com} & \email{grauman@cs.utexas.edu}\\
\affmark[1]The University of Texas at Austin & \affmark[2]Facebook AI Research
\end{tabular}
}

\maketitle

\begin{abstract}
The evolution of clothing styles and their migration across the world is intriguing, yet difficult to describe quantitatively.
We propose to discover and quantify fashion influences from catalog and social media photos.
We explore fashion influence along two channels: geolocation and fashion brands.
We introduce an approach that detects which of these entities influence which other entities in terms of propagating their styles.
We then leverage the discovered influence patterns to inform a novel forecasting model that predicts the future popularity of any given style within any given city or brand.
To demonstrate our idea, we leverage public large-scale datasets of 7.7M Instagram photos from 44 major world cities (where styles are worn with variable frequency) as well as 41K Amazon product photos (where styles are purchased with variable frequency).
Our model learns directly from the image data how styles move between locations and how certain brands affect each other's designs in a predictable way.
The discovered influence relationships reveal how both cities and brands exert and receive fashion influence for an array of visual styles inferred from the images.
Furthermore, the proposed forecasting model achieves state-of-the-art results for challenging style forecasting tasks.
Our results indicate the advantage of grounding visual style evolution both spatially and temporally, and for the first time, they quantify the propagation of inter-brand and inter-city influences. Project page:~\url{https://www.cs.utexas.edu/~ziad/influence_from_photos.html}
\end{abstract}

\begin{textblock*}{\textwidth}(0cm,-20cm)
\centering
To appear in the IEEE Transactions on Multimedia, 2020.
\end{textblock*}

\section{Introduction}

`The influence of Paris, for instance, is now minimal. Yet a lot is written about Paris fashion.' ---Geoffrey Beene
\vspace{0.1in}

The clothes people wear are a function of personal factors like comfort, taste, and occasion---but also wider and more subtle influences from the world around them, like changing social norms, art, the political climate, celebrities and style icons, the weather, and the mood of the city in which they live.
Fashion itself is an evolving phenomenon because of these changing influences.
What gets worn continues to change, in ways fast, slow, and sometimes cyclical.

Pinpointing the influences in fashion, however, is non-trivial.
To what extent did the runway styles in Paris last year affect what U.S. consumers wore this year?
How much did the designs by J. Crew influence those created six months later by Everlane, and vice versa?
How long does it take for certain trends favored in New York City to migrate to Austin, if they do at all?
And how did the infamous cerulean sweater worn by the protagonist in the movie \emph{The Devil Wears Prada} make its way into her closet?\footnote{The Devil Wears Prada: Cerulean influence~\url{https://bit.ly/3dBAQ5W}}

To quantitatively answer such questions would be valuable to both social science and the fashion industry, yet it remains challenging.
Clothing sales records or social media ``likes" offer some signal about how tastes are shifting, but they are indirect and do not reveal the sources of influence.

\begin{figure*}[t]
\centering
    \includegraphics[width=0.9\linewidth]{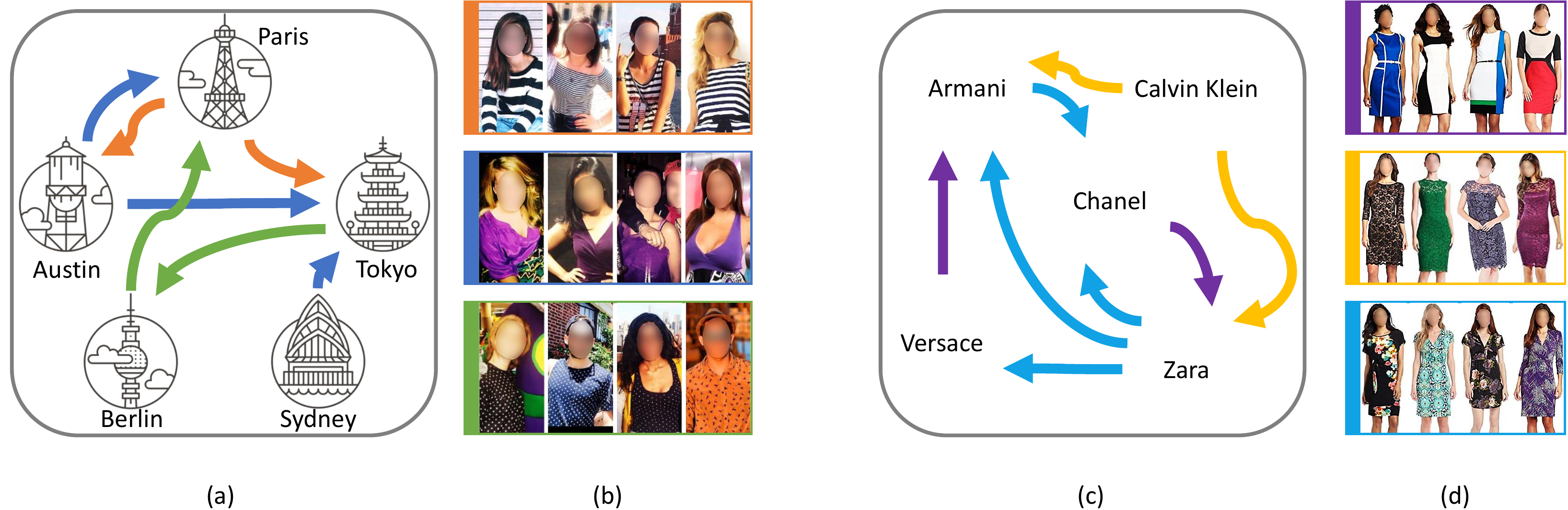}
\vspace*{-0.2cm}
\caption[]{Fashion styles propagate according to certain patterns of influence around the world.
Our idea is to learn visual styles from large-scale (b) social media photos or (d) catalog product images, and then discover style influence relations among (a) major cities worldwide or (c) fashion brands.
Our model leverages these relations to accurately forecast future fashion trends.  
City icons are attributed to \url{https://bit.ly/2WDyFqB}
}
\vspace*{-0.3cm}
\label{fig:intro}
\end{figure*}

We contend that images are exactly the right data to answer such questions.
Unlike vendors' non-visual meta-data or hype from haute couture designers,
photos of what people are wearing or buying in their daily lives provide a unfiltered glimpse of current clothing styles ``on the ground".
Our idea is to discover fashion influence patterns in community photos (\eg Instagram) and catalog photos (\eg Amazon), and leverage those patterns to forecast future style trends.
We explore fashion influence along two channels--cities and fashion brands---and forecast the popularity of a given style conditioned on the place in the world it will be worn or the brand from which it will be purchased.  See~\figref{fig:intro}.
Specifically, we aim to discover \emph{which cities influence which other cities} and \emph{which brands influence which other brands} in terms of propagating their clothing styles, and \emph{with what time delay}.

To this end, we introduce a unified approach to discover style influences from photos.
First, we extract a vocabulary of visual styles from unlabeled, timestamped photos.
We consider two sources of photos: geolocated social media photos from around the world, and catalog product photos from an online retailer.
Each style is a mixture of detected visual attributes.
For example, one style may capture short floral dresses in bright colors (\figref{fig:intro}d). 
Next, we record the past trajectories of each style's popularity, \ie the frequency with which it is seen in the photos (or purchased from the catalog) over time.

To build a predictive model, we identify two key properties of an influencer---time precedence and novelty---and use a statistical measure that captures these properties to calculate the degree of influence between cities or brands as well as the influences among the various styles themselves.
Next, we introduce a neural forecasting model that exploits the influence relationships discovered from photos to better anticipate future popular styles in any given location or brand.
Finally, we propose a novel coherence loss to train our model to reconcile the local predictions with the global trend of a style for consistent forecasts.
We demonstrate our approach on two large-scale datasets: GeoStyle~\cite{mall2019geostyle}, which is comprised of everyday photos of people with a wide coverage of geographic locations, and AmazonBrands~\cite{mcauley2015image}, which is comprised of catalog photos of clothing.
We gauge popularity for the visual styles using the frequency the various garments are worn or purchased, respectively.

Our work is the first to model and quantify fashion influence relations between cities and brands from real-world images.
Our results shed light on the spatio-temporal migration of fashion trends across the world---revealing which cities are exerting and receiving more influence on others, which most affect global trends, which contribute to the prominence of a given style, and how a city's degree of influence has itself changed over time.
Our findings hint at how computer vision can help democratize our understanding of fashion influence, sometimes challenging common perceptions about what parts of the world are driving fashion (consistent with designer Geoffrey Beene's quote above).  
Furthermore, our results examining influence between fashion brands exposes the latent effects that one line of clothing can have on another, as well as brands that are relatively impervious to the influence of any others.
In addition, we demonstrate that by incorporating influence, the proposed forecasting model yields state-of-the-art accuracy for predicting the future popularity of styles.
Unlike prior work that learns trends with a monolithic worldwide model~\cite{Al-Halah2017} or independent per city models~\cite{mall2019geostyle}, our influence-based predictions catch the temporal dependencies between when different cities or brands will see a style climb or dip, producing more accurate forecasts.

\section{Related Work}
\label{sec:related_work}

Visual fashion analysis, with its challenging vision problems and  direct impact on our social and financial life, presents an attractive domain for vision research.
In recent years, many aspects of fashion have been addressed in the computer vision literature, ranging from learning fashion attributes~\cite{berg2010automatic,bossard2012apparel,chen2012describing,Chen2015,liu2016}, landmark detection~\cite{Wang_2018_CVPR,Yu_2019_CVPR}, cross-domain fashion retrieval~\cite{liu2012street,huang2015cross,hadi2015buy,liang2016retreival,Zhao_2017_CVPR,Kuang_2019_ICCV}, body shape and size based fashion suggestions~\cite{misra2018decomposing,hidayati2018dress,kimberly-cvpr2020,hidayati2020style}, virtual try-on~\cite{wang2018toward,liu2019styletransfer,Dong_2019_ICCV}, clothing recommendation~\cite{liu2012hi,mcauley2015image,zhang2017recommendation,hsiao2019fashion++}, inferring social cues from people's clothes~\cite{song2011predicting,murillo2012urban,kwak2013bikers}, outfit compatibility~\cite{li2017outfit,hsiao2018creating,Yu_2019_ICCV,Han_2019_ICCV}, visual brand analysis~\cite{kim2013discovering,hadi2018brand}, and discovering fashion styles~\cite{kiapour2014hipster,veit2015learning,hsiao2017latent-look,Al-Halah2017}.
Our work opens a new avenue for visual fashion understanding:
modeling influence relations in fashion directly from images.

\parbf{Statistics of styles}
Analyzing styles' popularity in the past gives a window on people's preferences in fashion.
Prior work considers how the frequency of attributes (\eg floral, neon) changed over time~\cite{Vittayakorn2015,hidayati2014fashion}, and how trends in (non-visual) clothing meta-data changed for the two cities Manila and Los Angeles~\cite{Simo-Serra2015}.
Qualitative studies suggest how 
recommendation models can account for past temporal changes of fashion~\cite{He2016} or what cities exhibit strong style similarities~\cite{kataoka2019ten}.
However, all this prior work analyzes style popularity in an ``after the fact'' manner, and looks only qualitatively at past changes in style trends.
We propose to go beyond this historical perspective to forecast future changes in styles' popularity, and we provide supporting quantitative evaluation.

\parbf{Trend forecasting}
Most previous methods do not leverage visual information, instead focusing on demand forecasting of fashion based on sales records of clothing items (\eg~\cite{ren2014fashion,kaya2014fuzzy,thomassey2014sales,banica2014neural,ren2017comparative}).
Only limited prior work explores forecasting visual styles into the future~\cite{Al-Halah2017,mall2019geostyle}.
The FashionForward model~\cite{Al-Halah2017} uses fashion styles learned from Amazon product images to train an exponential smoothing model for forecasting, treating the products' transaction history (purchases)  as a proxy for style popularity.
Similarly, the GeoStyle project~\cite{mall2019geostyle} uses a seasonal forecasting model to predict changes in style trends per city. 
Both prior models assume that style trends in different cities are independent from one another and can be modeled monolithically~\cite{Al-Halah2017} or in isolation~\cite{mall2019geostyle}.
In contrast, we introduce a novel model that accounts for influence patterns discovered across different cities (or brands).
Our concept of fashion influence discovery is new, and our resulting forecasting model outperforms the state of the art.

\parbf{Influence modeling}
To our knowledge, no previous work tackles influence modeling in the visual fashion domain.
The closest study
looks at the correlation among attributes popular in New York fashion shows and those attributes seen in street photos, as a surrogate for fashion shows' impact on people's clothing; however, no influence or forecasting model is developed~\cite{Chen2015}.  Outside the fashion domain, models for influence are developed for connecting text in news articles~\cite{shahaf2010connecting}, linking video subshots for summarization~\cite{lu2013story}, or analyzing intellectual links between major AI conferences from their papers~\cite{chen2018modeling}.

In a prior conference paper, we introduced the first influence model based on visual fashion trends that is grounded by the style forecasting task~\cite{Al-Halah2020}.
Our previous model captures influence relations between major cities around the world, and we showed that it discovers interesting influence patterns in fashion that go beyond simple correlations.
Building on our earlier findings, this article expands both the model and experiments.
In particular, we extend our model to go beyond cities' influences to capture how styles themselves influence each other.
We propose a model that leverages both cities' and styles' influence relations, and we show it yields higher forecasting accuracy.
Furthermore, we provide an in-depth analysis of the relations discovered by our model and how the measured influence correlates with fashion experts and public opinion on what is fashionable.
Finally, we generalize our model to capture not only geographic influences but also influences among fashion \emph{brands}.  
We demonstrate that our approach captures how major fashion brands influence each other with a new set of experiments on another public dataset, AmazonBrands.

\section{Fashion Influence Model}
\label{sec:approach}

We propose an approach to model influence relations in fashion based on visual data.
We consider two domains of influence:
a) \emph{location influence}, where we quantify influence relations between different locations in the world (\eg cities) on what people wear in their everyday life;
b) \emph{brand influence}, where we capture the influence relations between fashion brands on what clothing items they sell.
Both locations and brands can be seen as equivalent concepts for our influence modeling since both can be represented with a distribution over visual fashion styles.
For brevity, we refer to a location or a brand as a \emph{fashion unit} in the following.

Starting with images of fashion garments, 1) we learn a visual style model that captures the fine-grained properties common among the garments (\secref{sec:app_style});
then 2) we construct style popularity trajectories by leveraging images' temporal and fashion unit meta information (\secref{sec:app_traj});
3) we model the influence relations between different units (\secref{sec:app_influence_unit}) and styles (\secref{sec:app_influence_style}). 
Finally, 4) we introduce a forecasting model that utilizes the learned influence relations together with a coherence loss (\secref{sec:app_forecast}) for consistent and accurate predictions of future style trends (\secref{sec:app_influence_full}).

\subsection{Visual Fashion Styles}
\label{sec:app_style}

Our model captures the fashion influence among different fashion units---\ie either locations or brands.
We begin by discovering a set of visual fashion \emph{styles} from images of people's garments.
Such images could be of people in everyday life and from around the world collected from photo-sharing social media platforms (\eg a public dataset of Instagram images) or catalog images of garments from an online retailer platform (\eg a public dataset of Amazon images).

Let $X=\{x_i\}^N$ be a set of clothing images.
We first learn a semantic representation that captures the elementary fashion attributes like colors (\eg cyan, green), patterns (\eg stripes, dots), shape (\eg v\-neck, sleeveless) and garment type (\eg shirt, sunglasses).
Given a fashion attribute model $f_a(\cdot)$ trained on a set of disjoint labeled images, we can then represent each image in $X$ with $\mathbf{a}_i=f_a(x_i)$, where $\mathbf{a}\in\mathbb{R}^M$ is a vector of $M$ visual attribute probabilities.
We train a convolutional neural network (CNN) for $f_a(\cdot)$. 

Next, we learn a set of fashion styles $S=\{S^k\}^K$ that capture distinctive attribute combinations.
Hence, given an image of a new garment $x_i$, the style model $f_s(\cdot)$ can predict the probabilities of that garment to be from each of the learned styles $\mathbf{s}_i=f_s(\mathbf{a}_i)$.
For example, the model $f_s(\cdot)$ can be realized using a Gaussian mixture model with $K$ components (see \secref{sec:eval_styles} for details on $f_s$ and $f_a$). The number of styles $K$  is selected following~\cite{Al-Halah2017,mall2019geostyle} to capture coherent visual appearance of a mid-level granularity.\footnote{Prior work~\cite{Al-Halah2017}
analyzes the impact of $K$ on forecasting: as $K$ increases the styles become less coherent and the noise in the trajectories increases since these styles have less supportive samples across time.}

\begin{figure}[t]
\centering
    \includegraphics[width=1.\linewidth]{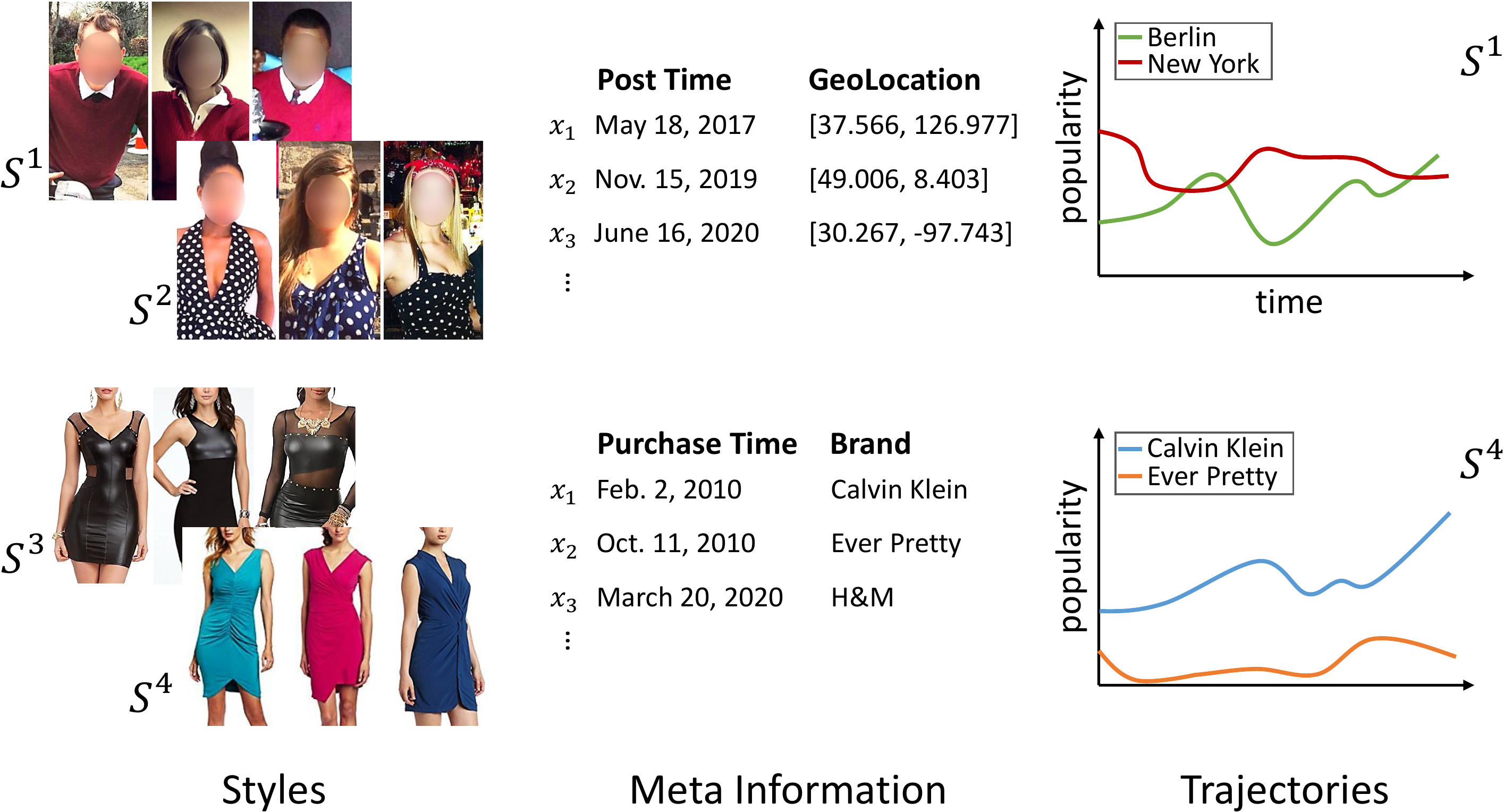}
\caption{Style trajectories.
    First, we learn a set of fashion styles from images (left).
    The images have meta information in terms of either timestamps and geolocations from social media (center top), or purchase time and brand labels from commercial catalog data (center bottom).
    Then based on the images' meta information, we measure the popularity of a style at a given city or for a given brand and a time period (\eg week) to build up its style popularity trajectory (right).
}
\vspace{-0.4cm}
\label{fig:styles_traj}
\end{figure}
 \figref{fig:styles_traj} (left) shows a set of fashion styles discovered by our style model from Instagram ($S^1$ and $S^2$) or Amazon product ($S^3$ and $S^4$) images.
Each style is a distribution over the visual attributes, capturing a particular ``look" that recurs within the data.
For example, in \figref{fig:styles_traj}, $S^2$ corresponds to a black and white spotted and sleeveless top whereas $S^4$ represents short dresses with uniform color and a v-neck.

\subsection{Style Trajectories}
\label{sec:app_traj}

We measure the popularity of a fashion style $S^i$ in the context of a certain fashion unit $U^j$ through the frequency of the style in the photos associated with that unit.
Specifically, let $Q$ be a set of transactions (\eg social media posts, purchases) such that each $q \in Q$ is a tuple of an image of fashion garment or outfit $x$, a time stamp $t$, and a fashion unit id $U$.
When we model city-city influences, the latter is the city; when we model brand-brand influences, it is the brand.
We construct a temporal trajectory $y^{ij}$ for each pair of style and unit $(S^i, U^j)$:
\begin{equation}\label{eq:traj}
    y^{ij}_{t} = \frac{1}{|U_{t}^{j}|} \sum_{x_k \in U_{t}^{j}} p(S^i|x_k),
\end{equation}
where $U_t^j$ is the set of images from fashion unit $U^j$ in the time window $t$, $p(S^i|x_k)$ is the probability of style $S^i$ given image $x_k$ based on our style model $f_s(\cdot)$, and $y^{ij}_t$ is the popularity of style $S^i$ in unit $U^j$ during time $t$.
The time step $t$ can have different temporal resolutions (\eg weeks, months) depending on the sought granularity of the trajectory.
Finally, by getting all values for $t={1,\dots, T}$ we build the trajectory $y^{ij}$.

\figref{fig:styles_traj} shows how these trajectories are constructed from social media posts and purchase logs via the large-scale datasets used in our experiments (detailed in Sec.~\ref{sec:eval_datasets}).
As depicted in the top panel, given the timestamps and geolocations of the photos, we quantize these properties to a meaningful temporal resolution (\eg weeks) and locations (\eg cities).
We then measure the popularity of each style in a city as in \eqref{eq:traj}, which captures the frequency with which people in city $j$ are seen wearing the style $i$ over time in the social media photos.
Similarly, when modeling fashion brands, we obtain style trajectories representing the popularity of style $i$ over time for brand $j$ in terms of the worldwide purchase rate of garments shown in the brand's catalog photos (\figref{fig:styles_traj} (bottom)).

Next, we describe our influence model that analyzes these trajectories to discover the influence patterns among the various fashion units.

\subsection{Fashion Unit Influence Modeling}
\label{sec:app_influence_unit}

We propose to ground fashion influence through style popularity forecasting.
This enables us to quantitatively evaluate influence using learned computational models based on real world data.

We say fashion unit $U^i$ influences unit $U^j$ in a given fashion style $S^n$ if our ability to accurately forecast the popularity of $S^n$ in $U^j$ significantly improves when taking into consideration the past popularity trend of $S^n$ in $U^i$, in addition to its past popularity trend in $U^j$.
In other words, past observations in $y^{ni}_{1\dots t}$ provide us with new insight on the future changes in $y^{nj}_{t+1\dots t+h}$ that are not available in $y^{nj}_{1\dots t}$.

\begin{figure}[t]
\centering
    \includegraphics[width=0.8\linewidth]{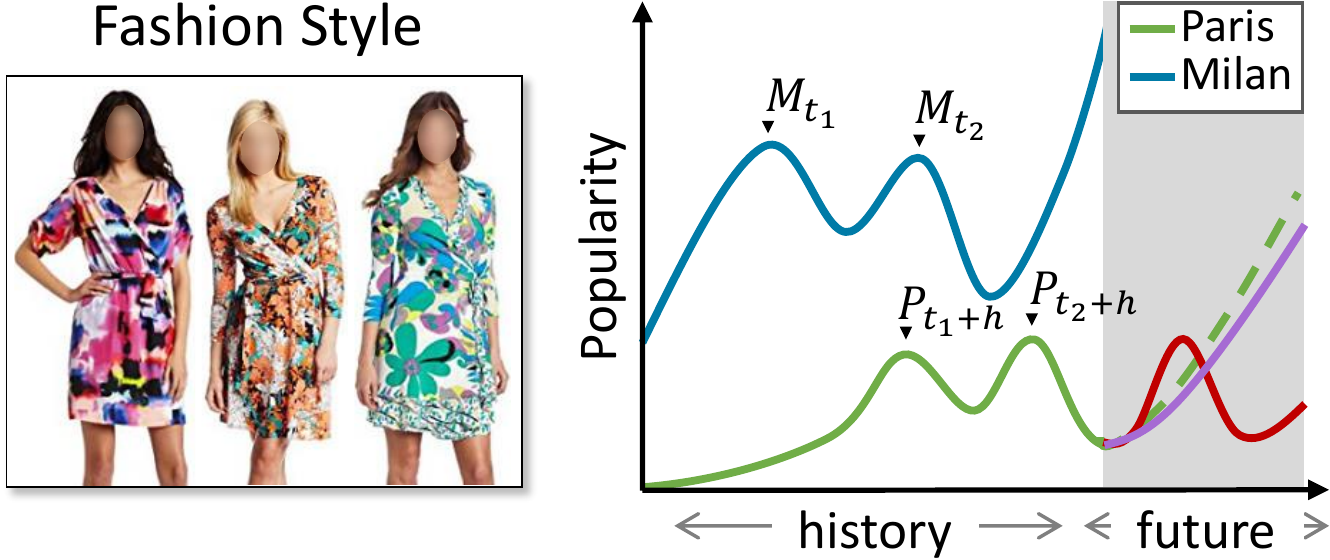}
\caption{
Influence modeling.
We propose to ground influence among fashion units with trend forecasting.
An influencer city (\eg Milan) has unique cues to accurately predict the future changes of a style trend (purple curve) for the influenced city (\eg Paris).
In contrast, forecasting without regard to influence can falter in the presence of complex trends (red curve).
We leverage this insight both for inter-city influences as well as inter-brand influences.
}
\vspace{-0.3cm}
\label{fig:infl_idea}
\end{figure}
 \figref{fig:infl_idea} demonstrates our idea for the case where the fashion units are cities.
We see that the style trend in \emph{Milan} foreshadows the changes in \emph{Paris} with a time lag of $h$.
By taking into consideration this influence relation, we can accurately predict the future changes in \emph{Paris} (purple curve).
On the other hand, modeling the style popularity for \emph{Paris} in isolation leads to significant forecasting error (red curve) since previous observations of style changes for this particular style in \emph{Paris} (\ie in the absence of the influencer) do not provide any cues for the sudden increase in the style's popularity.

We identify two main properties that an influencer $U^i$ should possess:
1) time precedence, that is the influencer unit's changes happen before the observed impact on the influenced unit and 2) novelty, that is the influencer unit has novel past information not observed in the history of the influenced unit.

A naive approach to capture such relations is to use a multivariate model to learn to predict $y^{nj}_t$ by feeding it all available information from the other units.
However, this approach does not satisfy the second property for an influencer since it does not constrain the influencer to have novel information that is not present in the influenced entity.
Instead, we capture our fashion influence relations using the Granger causality test~\cite{granger1969}.
The test determines that a time series $y^1$ Granger causes a time series $y^2$ if, while taking into account the past values of $y^2_{1\dots t}$, the past values of $y^1_{1\dots t}$ still have statistically significant impact on predicting $y^2_{t+1}$.
The test proceeds by modeling $y^2$ with an autoregressor of order $d$:
\begin{equation}\label{eq:ar_1}
    y^2_t = \phi_0 + \sum_{k=1}^d \phi_k y^2_{t-k} + \sigma_t,
\end{equation}
where $\sigma_t$ is an error term and $\phi_k$ contains the regression coefficients.
Then the autoregressor of $y^2$ is extended with lagged values of $y^1$ such that:
\begin{equation}\label{eq:ar_2}
    y^2_t = \phi_0 + \sum_{k=1}^d \phi_k y^2_{t-k} + \sum_{l=m}^q \psi_l y^1_{t-l} + \sigma_t,
\end{equation}
where the third term on the righthand side is a regressor on the lags from $y^1$.
If these extended lags from $y^1$ do add significant explanatory power to $y^2_t$, \ie the forecast accuracy of $y^2$ is significantly better ($p<0.05$) according to a regression metric (mean squared error), then $y^1$ Granger causes $y^2$.

We estimate the influence relations across all units' trajectories for each fashion style $S^i$.
In experiments we consider lags ranging from $1$ to $8$ temporal steps, meaning up to two months.
In this way, we establish the influence relations among units and at which lag this influence occurs.
Note that an influence relationship discovered for two cities or two brands indicates that their trends are linked in some predictable way.
In particular, a Granger causality tie between two units does not mean that the influenced unit necessarily repeats the shape of the influencer's trajectory at a later time, only that the style trajectory for one unit can be more accurately forecasted by knowing the past trajectory of the other.

\begin{figure}[t]
\centering
    \includegraphics[width=0.9\linewidth]{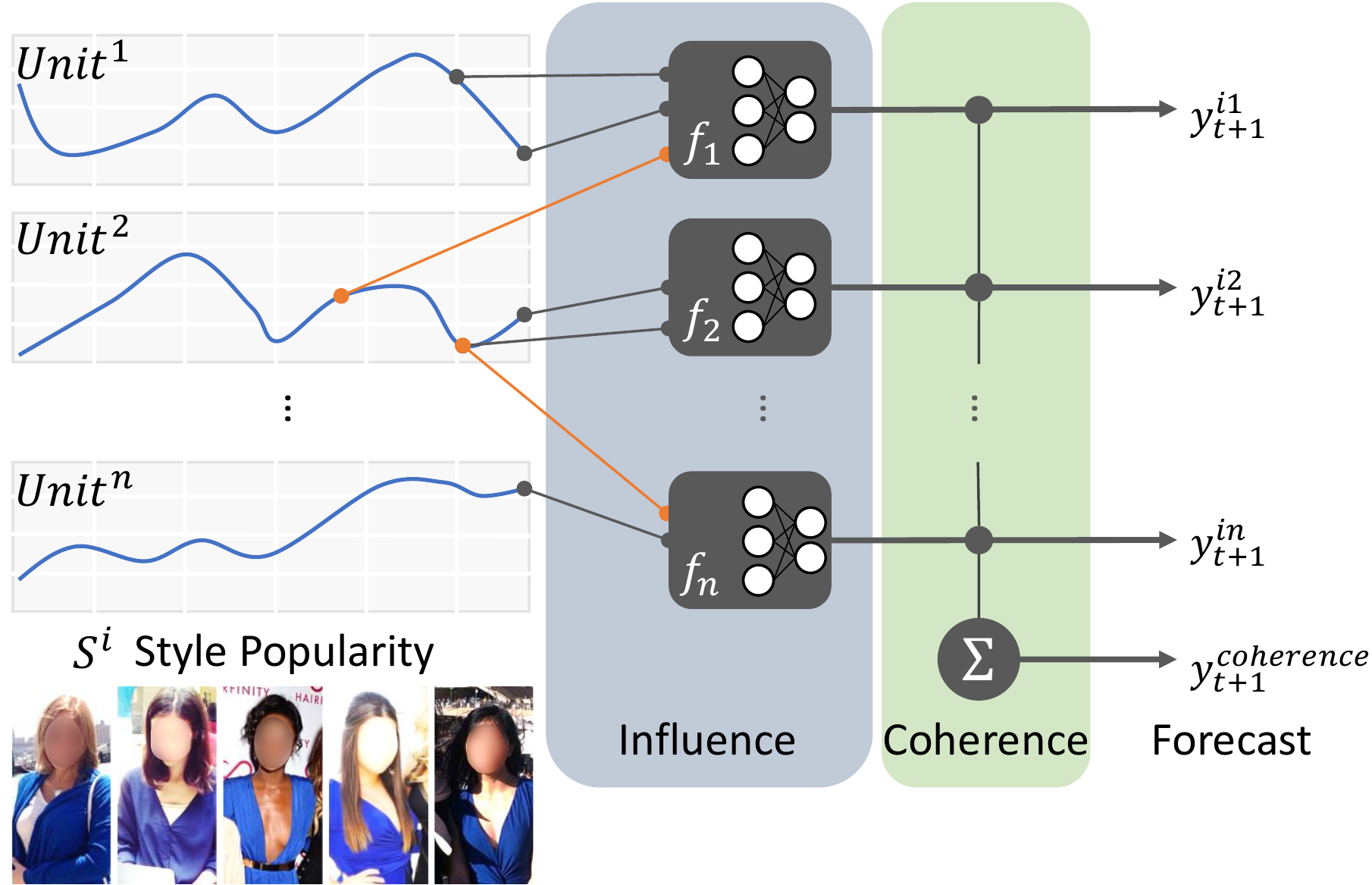}
\caption{Influence coherent forecaster.
    Our model captures influence relations between fashion units for a given fashion style (orange connections) and uses them to predict future changes in the style popularity for each unit.
    Additionally, our model regularizes the forecasts to be coherent with the global trend of the style observed across all units.
    }
\vspace{-0.3cm}
\label{fig:model}
\end{figure}
 \subsection{Coherent Style Forecaster}
\label{sec:app_forecast}

After we estimate the influence relations across the fashion units, we build a forecaster for each trajectory $y^{ij}$ such that:
\begin{equation}\label{eq:infl_forecast}
    \tilde{{y}}^{ij}_{t+1} = f_{unit}(L(y^{ij}_t),I(y^{ij}_t) | \theta),
\end{equation}
where $I(y^{ij}_t)$ is the set of lags from the influencer of $y^{ij}$ relative to time step $t$ as determined in the previous section, and $L(y^{ij}_t)$ are the lags from $y^{ij}$'s own style popularity trajectory.
We model $f(\cdot)$ using a multilayer perceptron (MLP) and estimate the parameters $\theta$ by minimizing the mean squared error loss:
\begin{equation}\label{eq:loss_forecast}
    \mathcal{L}_{forecast} = \sum_t (y^{ij}_{t+1} - f(L(y^{ij}_t),I(y^{ij}_t) | \theta))^2,
\end{equation}
where $y^{ij}_{t+1}$ is the ground truth value of $y^{ij}$ at time $t+1$.

Our forecast model in its previous form does not impose any constraints on the forecasted values in relation to each other.
However, while we are forecasting the style popularity for each individual unit given the influence from the others, the forecasted popularities ($y^{i1}_{t}, y^{i2}_{t} \dots y^{in}_{t}$) are still for a common fashion style $S^i$ that by itself exhibits a worldwide trend across all locations or brands.

We propose to reconcile the base forecasts produced at each location through a \emph{coherence loss} that captures the global trend.
For all forecasts $\hat{y}^{ij}_{t+1}$ for a fashion style $S^i$ and across all units $U^j\in U$, we constrain the distribution mean of the predicted values to match the distribution mean of the ground truth values:
\begin{equation}\label{eq:loss_coherence}
    \mathcal{L}_{coherence} = \frac{1}{|U|} (\sum_k y^{ik}_{t+1} - \sum_k \hat{y}^{ik}_{t+1}).
\end{equation}

The coherence loss, in addition to capturing the global trend of $S^i$, helps in combating noise at the trajectory level of each unit through regularizing the mean distribution of all forecasts.
The model is trained with the combined forecast and coherence losses:
\begin{equation}\label{eq:loss}
    \mathcal{L} = \mathcal{L}_{forecast} + \lambda \mathcal{L}_{coherence}.
\end{equation}

\figref{fig:model} illustrates our model. 
For a style $S^i$, we model its popularity trajectory for each fashion unit with a neural network of two layers and sigmoid non-linearity.
The input of the network is defined by the lags from its own trajectory (shown in black) and any other influential lags from other units discovered by the previous step (\secref{sec:app_influence_unit}), which are shown in orange.
Furthermore, the output of all local forecasters is further regularized to be coherent with the overall observed trend of $S^i$ using our coherence loss.

\subsection{Style Influence Modeling}
\label{sec:app_influence_style}

\begin{figure}[t]
\centering
    \includegraphics[width=0.9\linewidth]{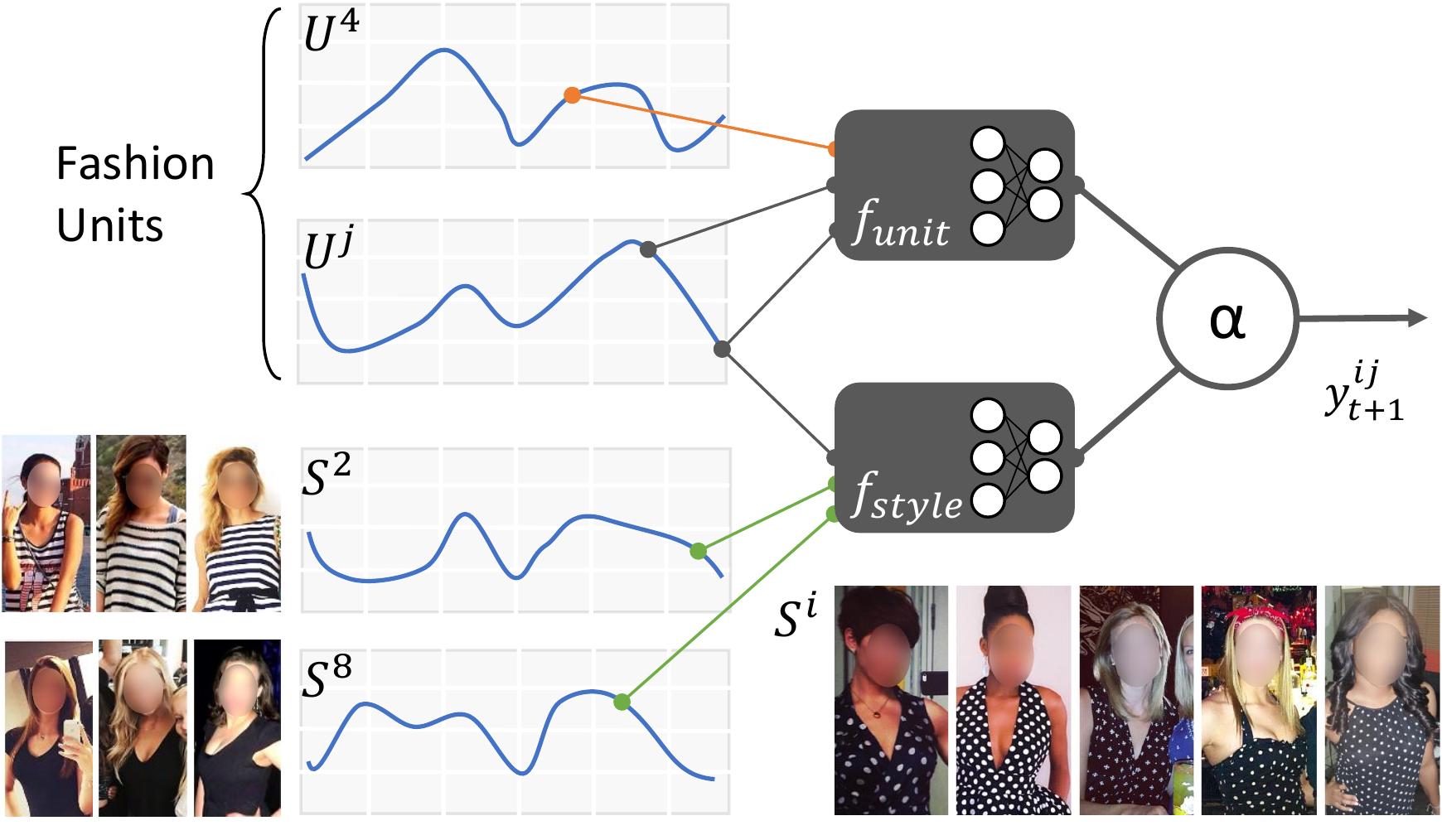}
\caption{
    Our influence-based model forecasts a style popularity by jointly modeling two types of influence: unit (orange) and style (green) influences. 
    }
\vspace{-0.3cm}
\label{fig:model_ens}
\end{figure}
 
We have so far modeled how fashion units (cities or fashion brands) influence each other for a given fashion style.
However, another type of influence to consider is how fashion styles impact each other's popularity.
For example, a spike in popularity of solid purple v-necks could foreshadow a spike in v-necks with more complicated textures and colors, or an upward tick in more conservative styles may foreshadow a downward trend in less conservative styles.
To capture these links within our model, we represent how a style $S^i$ influences another style $S^j$ in a unit $U^n$ in a similar manner as in \secref{sec:app_influence_unit} by simply swapping styles and units.

\subsection{Influence-based Forecasting}
\label{sec:app_influence_full}

\begin{figure}[t]
\centering
    \includegraphics[width=1.0\linewidth]{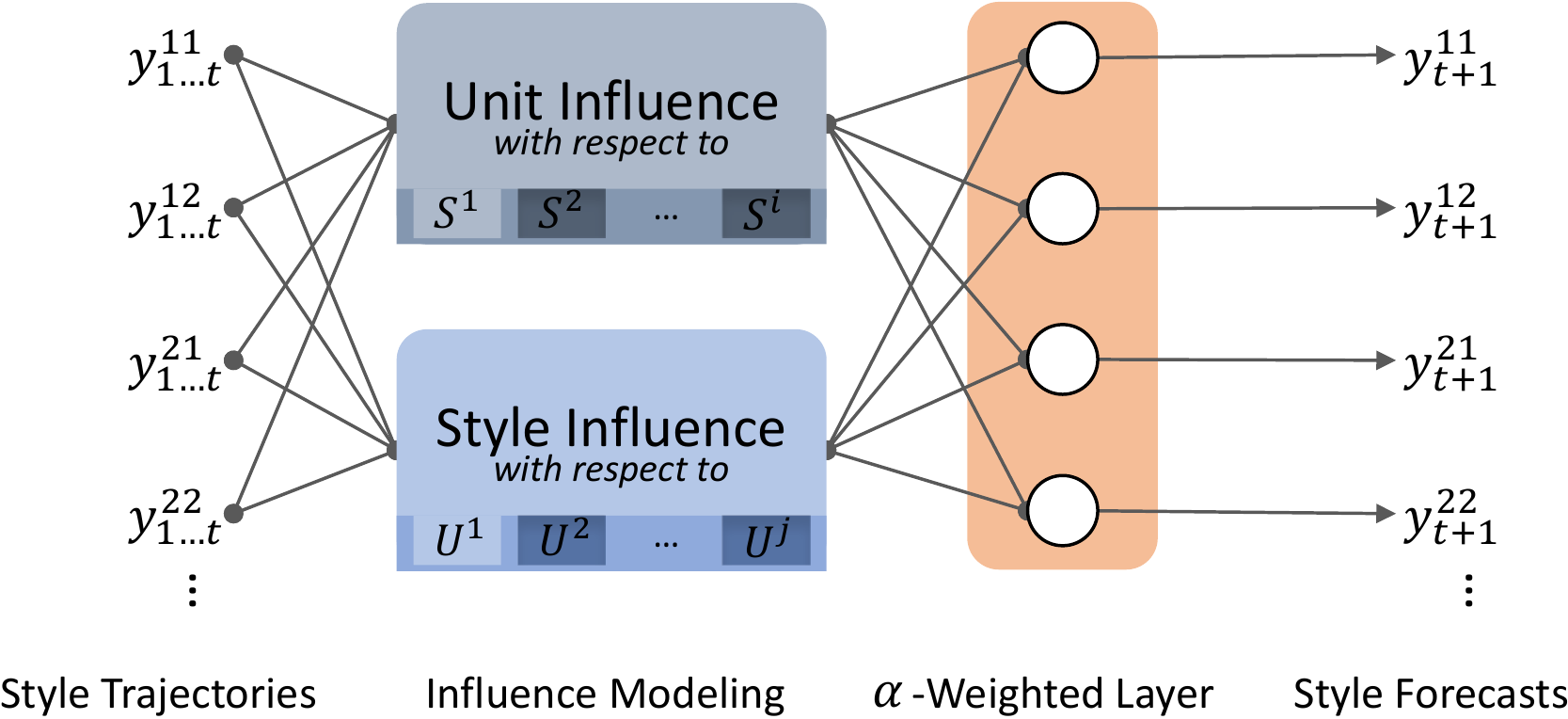}
\caption{
    Model overview.
    For each style trajectory, our model captures the influence of other fashion units and styles on the input trajectories using the Unit and Style Influence modules. Then we use an $\alpha$-weighted layer to combine the predictions of these modules and forecast the future trend of the styles.  This figure unifies the model components depicted in Figures~\ref{fig:model} and~\ref{fig:model_ens}.    }
\vspace{-0.3cm}
\label{fig:model_overview}
\end{figure}
 
Our final model forecasts the popularity of a style based on both types of influence:
    \begin{eqnarray}\label{eq:icf}
        \hat{{y}}^{ij}_{t+1} = \alpha f_{style}(L(y_t^{ij}), I(y_t^{ij})|\theta_s) +\nonumber\\ ~~~~~~~~~~(1-\alpha)f_{unit}(L(y_t^{ij}), I(y_t^{ij}) | \theta_u),
    \end{eqnarray}
where
$f_{style}$ and $f_{unit}$ are instances of our influence-based coherent forecaster (\eqref{eq:infl_forecast}) that capture style and unit influence relations, respectively.
\figref{fig:model_ens} shows our joint model of both types of influence relations.
Each $f_{style}$ and $f_{unit}$ is trained for its respective loss in Eq.~\ref{eq:loss}.
The hyperparameter $\alpha$ is learned over the validation data to weight each type of influence contribution to the final forecast.
We model theses influence relations using \eqref{eq:icf} for each of the trajectories in our data with respect to all fashion units and styles, as shown in the model overview in~\figref{fig:model_overview}.

\begin{figure*}[!t]
\centering
\subfloat[GeoStyle]{
    \includegraphics[width=0.40\linewidth]{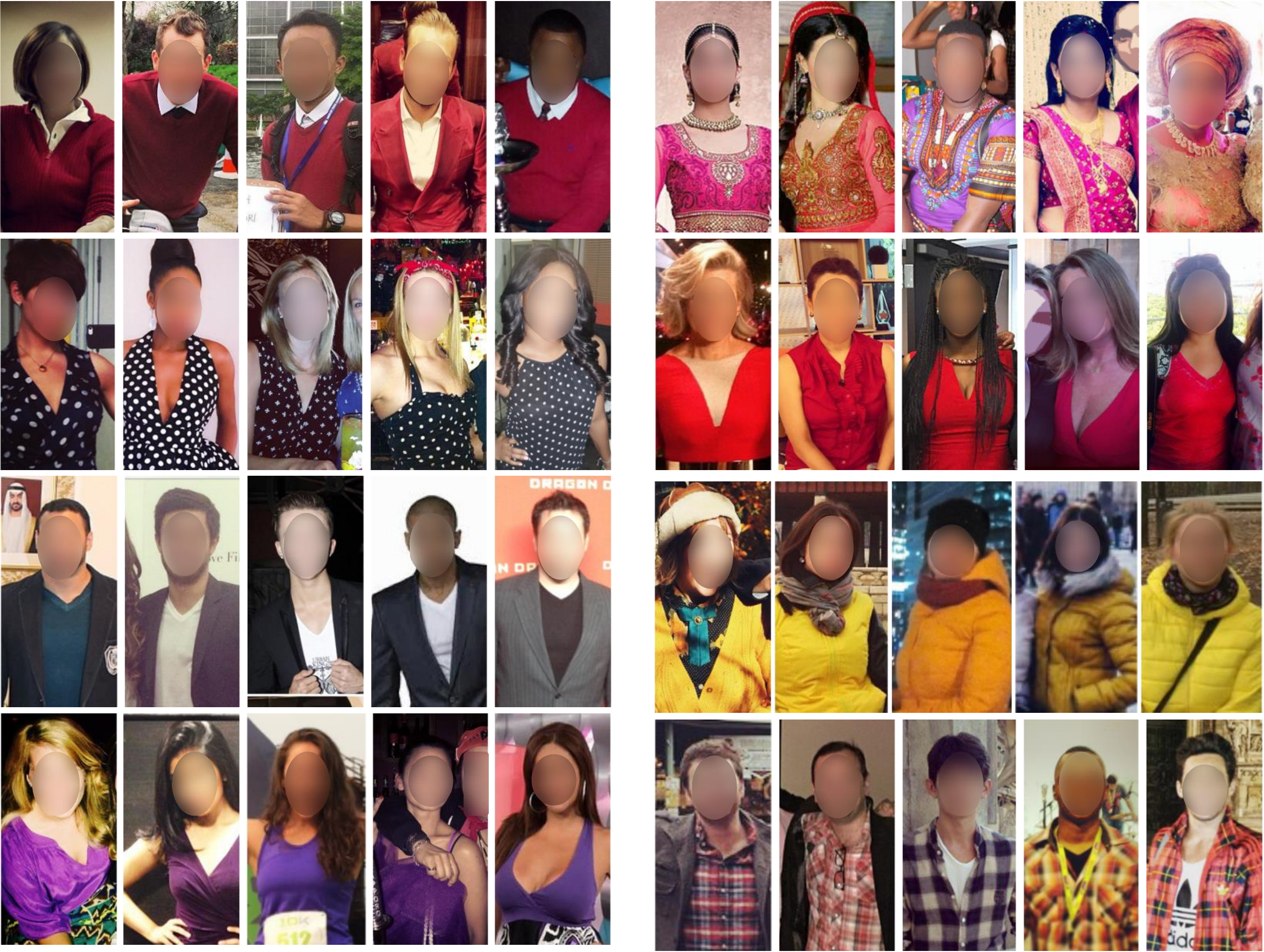}
    \label{fig:styles_examples_geo}
    }\hfil
\subfloat[AmazonBrands]{
    \includegraphics[trim={0.2cm 0cm 0.2cm 0cm} ,clip,width=0.40\linewidth]{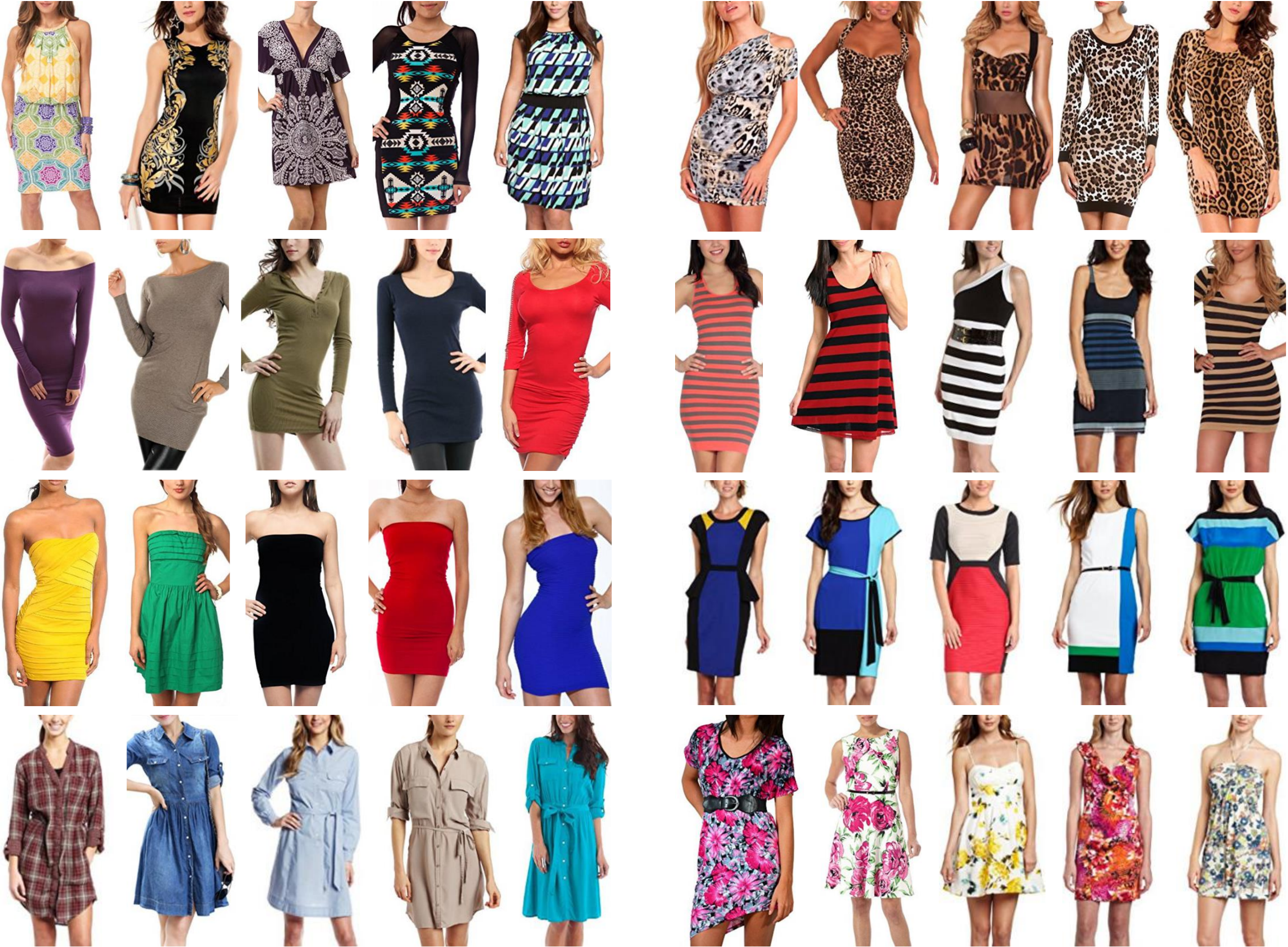}
    \label{fig:styles_examples_ama}
    }
\caption{
    Examples of the learned fashion styles from GeoStyle (a) and AmazonBrands (b) datasets.
    For each dataset, each row shows two example styles, with five sample images for each.
}
\vspace{-0.3cm}
\label{fig:styles_examples}
\end{figure*}
 \section{Evaluation}
\label{sec:eval}

In the following experiments, we demonstrate our model's ability to forecast fashion styles' popularity changes by utilizing discovered influence relations.
Furthermore, we analyze the influence patterns revealed by our model between major cities and brands, how they influence global fashion trends, and their influence dynamics trends through time.

\subsection{Datasets}
\label{sec:eval_datasets}

We evaluate our approach on two datasets:

\paragraph{GeoStyle~\cite{mall2019geostyle}}
This dataset, collected by researchers at Cornell University, extends the StreetStyle dataset~\cite{matzen2017streetstyle}. It is based on public Instagram photos showing people from $44$ major cities from around the world.
Since a photo may contain multiple persons and a cluttered background, the images are first preprocessed with a person detector to extract the regions of interest (see~\cite{matzen2017streetstyle} for details).
In total, the dataset has $7.7$ million images that span a time period from July 2013 until May 2016.

\paragraph{AmazonBrands}
This dataset is derived from the data collected by researchers at the University of California~\cite{mcauley2015image} from the Amazon website which contains images of garments along with their transaction history (purchases).
We use the \emph{dresses} subset~\cite{Al-Halah2017} which has $41K$ images.
We consider a list of the most famous fashion brands and match their names with the textual description associated with each sample in order to label each garment with its respective brand.
Finally, we select the top $10$ most frequent brands and their corresponding $4K$ images and $21K$ transactions to use in our experiments.
The samples span a time period from March 2012 until July 2014.  For this dataset, no person detection is needed since the catalog photos have white backgrounds with the fashion item centered.
We use both datasets for research purposes only.

These two datasets provide us with different perspectives on people's preferences in fashion.
While in GeoStyle we estimate a style's popularity based on its frequency in social media posts (\ie what you wear is what you like), in AmazonBrands the popularity is estimated based on purchase frequency (\ie what you buy is what you like).

However, both datasets, like any Internet photo dataset, have certain biases based on their respective data sources and sampling methods.
These biases may affect the type of styles considered and their measured popularity.
For example, images in GeoStyle are usually of young people in major cities since younger generations are more likely to upload photos to Instagram and from places with easy access to high Internet bandwidth~\cite{matzen2017streetstyle}.
On the other hand, while fashionability may be the main drive to select a given dress in AmazonBrands, other factors like price and brand loyalty could contribute to one's decision to buy as well.
Nonetheless, the GeoStyle dataset is the largest public fashion dataset with the most temporal and geographic coverage, providing a unique glimpse on people's fashion preferences in daily life around the globe.
Similarly, the AmazonBrands data gives us a unique insight on the influence dynamics among major fashion brands.

\subsection{Fashion Styles}
\label{sec:eval_styles}

A style is a combination of certain attributes describing materials, colors, cut, and other factors, such as \emph{V-neck, red, formal dress}.
Hence, we adopt an attribute-based representation of images for style learning.

For GeoStyle we use attribute predictions from~\cite{mall2019geostyle} to represent each photo with $M=46$ fashion attributes (\eg colors, patterns and garment types).
The attribute predictor is a multi-task CNN (GoogLeNet~\cite{szegedy2015going}) with separate heads to predict separate attributes.
This yields an $M$-dimensional vector of attribute probabilities per image.
Based on these, we learn $K=50$ fashion styles using a $M$-dimensional Gaussian mixture model: each style is a mode in the data capturing a distribution of attributes.
Then, we get the style probabilities of a garment by calculating the posterior probability of each mode in the mixture with respect to the inferred attributes' probabilities from the photo.

For AmazonBrands we train a ResNet-18 model \cite{he2016deep} to predict $M=1000$ fashion attributes learned from the DeepFashion dataset~\cite{liu2016}.
We learn fashion styles from these attributes using a non-negative tensor factorization (NMF) style model similar to~\cite{Al-Halah2017}.
That is, given a tensor representing the samples in the dataset by their attributes, we factorize it into two matrices using a set of learned latent variables.
Each of these latent variables represents a fashion style.
The NMF model uses a probabilistic formulation~\cite{hu2015scalable} and each discovered style captures a distribution of attributes in a similar fashion to the previous GMM model.
We find that for the high-dimensional attribute representation in AmazonBrands, NMF produces more coherent and diverse visual styles compared to the GMM model, which seemed to collapse on a few attribute dimensions for this data.
We learn $K=20$ styles, which was sufficient to cover the diverse styles in this dataset since AmazonBrands is focused on one kind of garment---dresses.

\figref{fig:styles_examples} shows examples of the fashion styles learned by our style models from the GeoStyle and AmazonBrands datasets.
We notice that in GeoStyle (\figref{fig:styles_examples_geo}) some of the learned fashion styles may reflect a season-related type of garment (\eg the yellow jacket and scarf style), or a local traditional or cultural clothing (\eg upper left row). 
However, many of the learned styles are common across different countries and cultures.
Differently, the styles learned in AmazonBrands (\figref{fig:styles_examples_ama}) do not suggest any clear seasonal variations.
They capture a coherent visual representation that can be described with signature attributes like the slim-fit leopard dresses (upper right row) or the casual long-sleeved short dresses (lower left row).
In both datasets, since each style is a mixture of attributes, it captures a recurring \emph{configuration} of properties, rather than one isolated property (\eg leopard-print plus tight fit; blazer paired with V-neck t-shirt).
Thus styles offer a coarser representation than individual clothing items, and they emerge bottom-up from the data.

\subsection{Style Trends Forecasting}
\label{sec:eval_forecast}

We evaluate how well our model produces accurate forecasts by leveraging the influence patterns, and we compare it to several baselines and existing methods that model trajectories in isolation~\cite{Al-Halah2017,mall2019geostyle}.

\subsubsection{Trajectories and data splits}
\label{sec:eval_forecast_traj_splits}
For GeoStyle, we infer the popularity trajectory of each style in each city based on its frequency in the images.
Additionally, to quantify the impact of possible seasonal yearly trends in fashion styles, we also consider forecasting the deseasonalized style trajectories.
To do so, we subtract the yearly seasonal lag from the trajectories.
The deseasonalized test is interesting because it requires methods to capture the more subtle visual trends not simply associated with the location's weather and annual events.
For AmazonBrands, we build the temporal trajectory of each style and brand based on its transaction frequency.

For both datasets, we use a temporal resolution of weeks.
We adopt the long-term forecasting data split
from \cite{mall2019geostyle}.
That is, we allocate the last $26$ weeks of each trajectory for testing, the previous $4$ for validation, and the rest for training.

\begin{table*}[t]
\setlength{\tabcolsep}{12pt}
\centering
\caption{Forecast errors of fashion style trajectories on the GeoStyle and the AmazonBrands datasets. } 
\label{tbl:forecast_styles}
\scalebox{0.9}{
\begin{tabular}{l c c c c c c}
\toprule
                    & \multicolumn{4}{c}{GeoStyle}                          & \multicolumn{2}{c}{AmazonBrands}\\
    & \multicolumn{2}{c}{Seasonal} & \multicolumn{2}{c}{Deseasonalized}     &       & \\
    Model          &  MAE      &    MAPE   &   MAE     &    MAPE       &   MAE     &    MAPE \\
\midrule
    \multicolumn{7}{l}{\textbf{Naive}}  \\
    Gaussian       &   0.1301   &   33.23   &   0.1222  &   26.08       &   0.1268  &   35.33        \\
    Seasonal       &   0.0925   &   22.64   &   0.1500  &   33.39       &   0.1329  &   36.48       \\
    Mean           &   0.0908   &   23.57   &   0.0847  &   18.97       &   0.0964  &   26.29        \\
    Last           &   0.0893   &   22.20   &   0.1053  &   23.08       &   0.1134  &   31.41        \\
    Drift          &   0.0956   &   23.65   &   0.1163  &   25.32       &   0.1207  &   33.53        \\
\midrule
    \multicolumn{7}{l}{\textbf{Per-Trajectory Models}} \\
    AR             &   0.0846   &   21.88   &   0.0846  &   18.95       &   0.0934  &   25.61   \\ 
    ARIMA          &   0.0919   &   23.70   &   0.1033  &   22.70       &   0.1095  &   30.77   \\ 
    FashionForward~\cite{Al-Halah2017}
                   &   0.0779   &   19.76   &   0.0848  &   18.94       &   0.0930  &   25.56   \\
    GeoModel~\cite{mall2019geostyle}
                   &   0.0715   &   17.86   &   0.0916  &   20.31       &   0.1028  &   28.41    \\
\midrule
    \multicolumn{7}{l}{\textbf{Across-Trajectories Models}}\\
    VAR - All Units           &   0.0771   &   19.25    &   0.0929   &   20.41      &    0.0955   &     26.12      \\ 
    VAR - All Styles          &   0.0752   &   19.15    &   0.0939   &   20.78      &    0.0949   &     26.00      \\ 
\midrule
    Influence-based (ours)     &  \textbf{0.0688}   &   \textbf{17.13}     &  \textbf{0.0814}   &  \textbf{18.10} &  \textbf{0.0914}   &   \textbf{24.86} \\
\bottomrule
\end{tabular}
}
\end{table*}
 \subsubsection{Implementation details}
We set $\lambda=1$ for the coherence loss weight (see \eqref{eq:loss}) and optimize our neural influence model using Adam for stochastic gradient descent with a learning rate of $10^{-2}$ and $l_2$ weight regularization of $10^{-8}$.
We select the best model and $\alpha \in [0,1]$ from \eqref{eq:icf} that controls the contributions of unit and style influences based on the performance on a disjoint validation split.

We compare all models using the forecast accuracy captured with the mean absolute error $\mathrm{MAE} = \frac{1}{n}\sum_{t=1}^{n}|\hat{y}_t - y_t|$, which measures the absolute difference between the forecasted $\hat{y}_t$ and ground truth $y_t$ values, and the mean absolute percentage error $\mathrm{MAPE} = \frac{1}{n}\sum_{t=1}^{n}|\frac{\hat{y}_t - y_t}{y_t}| \times 100$, which measures the forecast error scaled by the ground truth values, following prior work~\cite{Al-Halah2017,mall2019geostyle}.
The models are evaluated on forecasting a long time horizon of $26$ temporal steps, meaning six months forward in time.

\subsubsection{Baselines and existing methods}
We arrange the baselines into three main groups:

\parbf{Naive models}: these models rely on basic statistical properties of the trajectory to produce a forecast.
We consider five variants of these baselines:
\begin{itemize}
    \item Gaussian: this model fits a Gaussian distribution based on the mean and standard deviation of the trajectory and forecasts by sampling from the distribution.
    \item Seasonal: this model forecasts the next step to be similar to the observed value one season before $y_{t+1} = y_{t-season}$.  We set a yearly season of 52 weeks.
    \item Mean: it forecasts the next step to be equal to the mean observed values $y_{t+1} = \textrm{mean}(y_1,\dots,y_t)$.
    \item Last: it uses the value at the last temporal step to forecast the next $y_{t+1} = y_t$.
    \item Drift: it forecasts the next steps along the line that fits the first and last observations.
\end{itemize}

\parbf{Per-Trajectory models}:
These models fit a separate parametric model trained on the history ``lags'' of each of the trajectories~\cite{box2015time}, without accounting for relationships between the units (locations or brands) or styles.
\begin{itemize}
    \item AR: a standard autoregression model.
    \item ARIMA: a standard autoregressive integrated moving average model.
    \item  FashionForward~\cite{Al-Halah2017}: an exponential decay model which forecasts based on a learned weighted average of the historical values.
    \item  GeoModel~\cite{mall2019geostyle}: a parametric seasonal forecaster which models a trajectory with a sinusoidal model and drift components.
\end{itemize}
To our knowledge, the two existing methods~\cite{Al-Halah2017,mall2019geostyle} represent the only prior approaches for style forecasting from images.
Further, unlike our approach, all of these models consider the popularity trajectories of the styles in isolation, \ie they do not take into consideration possible interactions among the units or styles.

\begin{table}[!t]
\setlength{\tabcolsep}{15pt}
\centering
\caption{Ablation study of our model on the GeoStyle deseasonalized style trajectories. } 
\label{tbl:forecast_ablation}
\scalebox{.9}{
\begin{tabular}{l c c}
\toprule
    Model                       &  MAE      &   MAPE \\
\midrule
    Ours (full model)            &   0.0814   &   18.10  \\
\midrule
    Style Influence Only        &   0.0825   &   18.32   \\
    City Influence Only         &   0.0824   &   18.29   \\ 
    w/o Influence               &   0.0859   &   19.24   \\ 
    w/o Influence \& Coherence  &   0.0942   &   20.62   \\ 
\bottomrule
\end{tabular}
}
\vspace{-0.3cm}
\end{table}

 \parbf{Across-Trajectories models:}
The next set of baselines does model dependencies between the units or styles.
To represent this group, we use the VAR model~\cite{box2015time}, which fits a parametric model trained on the trajectories of a style across all units (VAR - All Units) or the trajectories from one unit and across all styles (VAR - All Styles).
Such models assume a full and simultaneous interaction between all units or styles.

\subsubsection{Forecasting results}

\tblref{tbl:forecast_styles} shows the performance of all models when forecasting the styles' future popularity.
The proposed model outperforms all the naive, per-trajectory, and across-trajectories models, attaining the lowest forecast errors.
This shows the value of discovering influence for the quantitative forecasting task.
We notice that the styles' popularity trajectories in GeoStyle do have a strong seasonal component: seasonal models (like GeoModel~\cite{mall2019geostyle} and Seasonal) do well compared to non-seasonal ones (like AR and Drift), but still underperform our approach.
This ranking changes on the deseasonalized test of GeoStyle and on AmazonBrands, where models like FashionForward~\cite{Al-Halah2017} and AR outperform the seasonal ones.
Our model outperforms all competitors on both types of trajectories, which demonstrates the benefits of accounting for influence.
Moreover, we compare our model's predictions to the best ``non-influence'' per-trajectory or across-trajectories competitors on both datasets using a two-sided t-test with the null hypothesis that the expected forecast error is identical. We found that our model's improvements are statistically significant with $p<0.05$.

\begin{figure*}[!t]
\centering
\subfloat[European Cities]{
    \includegraphics[trim={1.5cm 3cm 1.5cm 1.5cm} ,clip,width=0.30\linewidth]{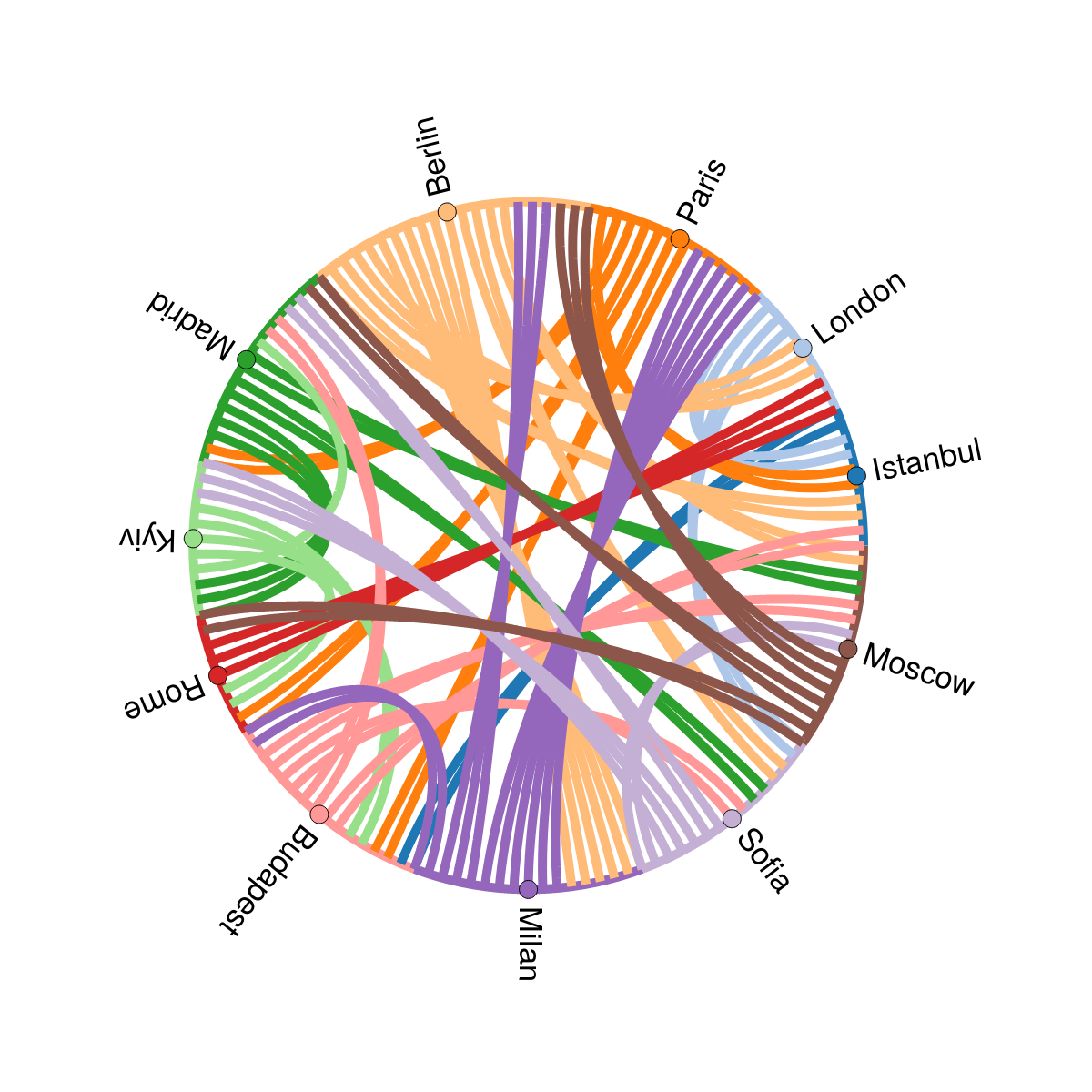}
    \label{fig:infl_cities_european}
    }
\subfloat[Asian Cities]{
    \includegraphics[trim={1.5cm 3cm 1.5cm 1.5cm} ,clip,width=0.30\linewidth]{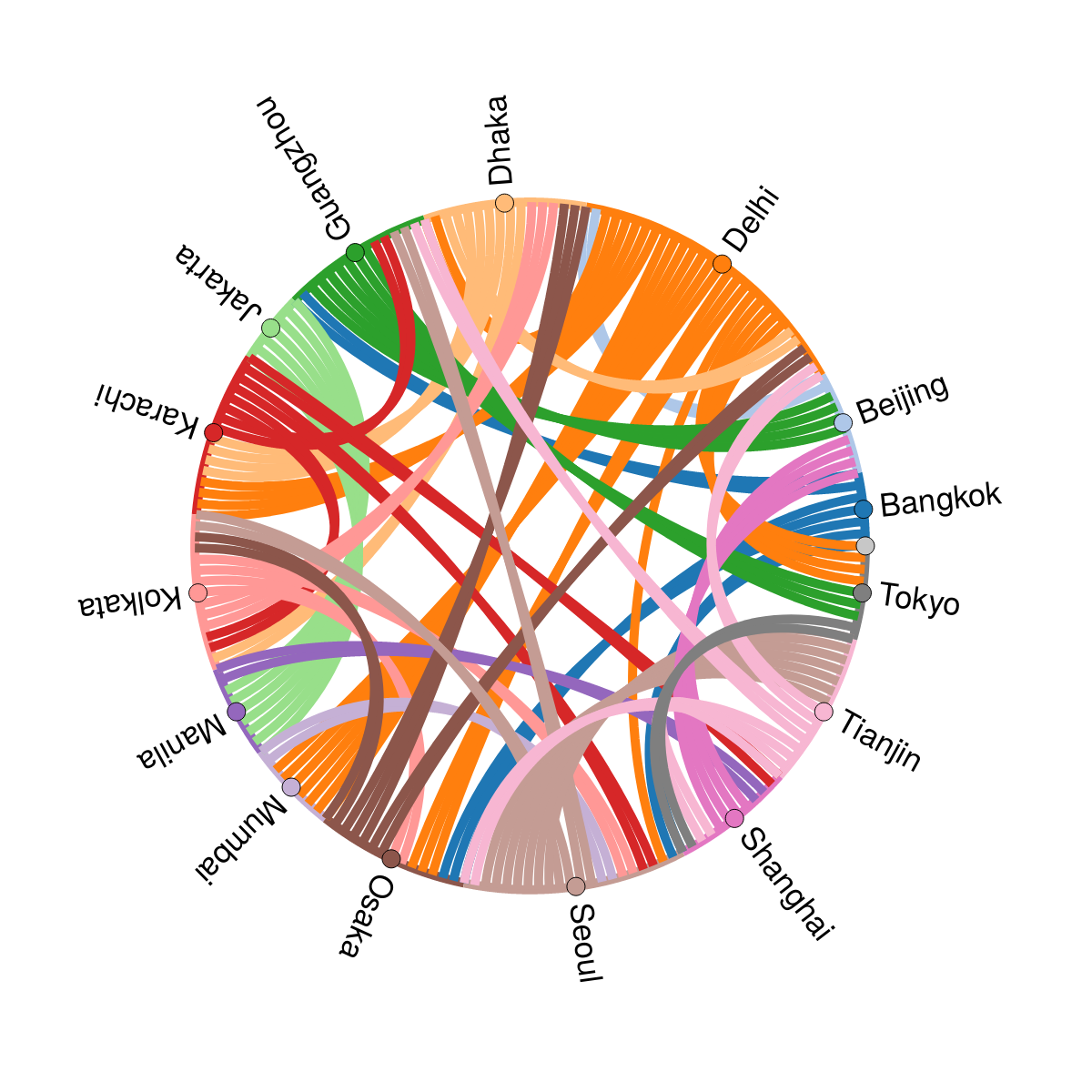}
    \label{fig:infl_cities_asian}
    }
\subfloat[American Cities]{
    \includegraphics[trim={1.5cm 3cm 1.5cm 1.5cm} ,clip,width=.30\linewidth]{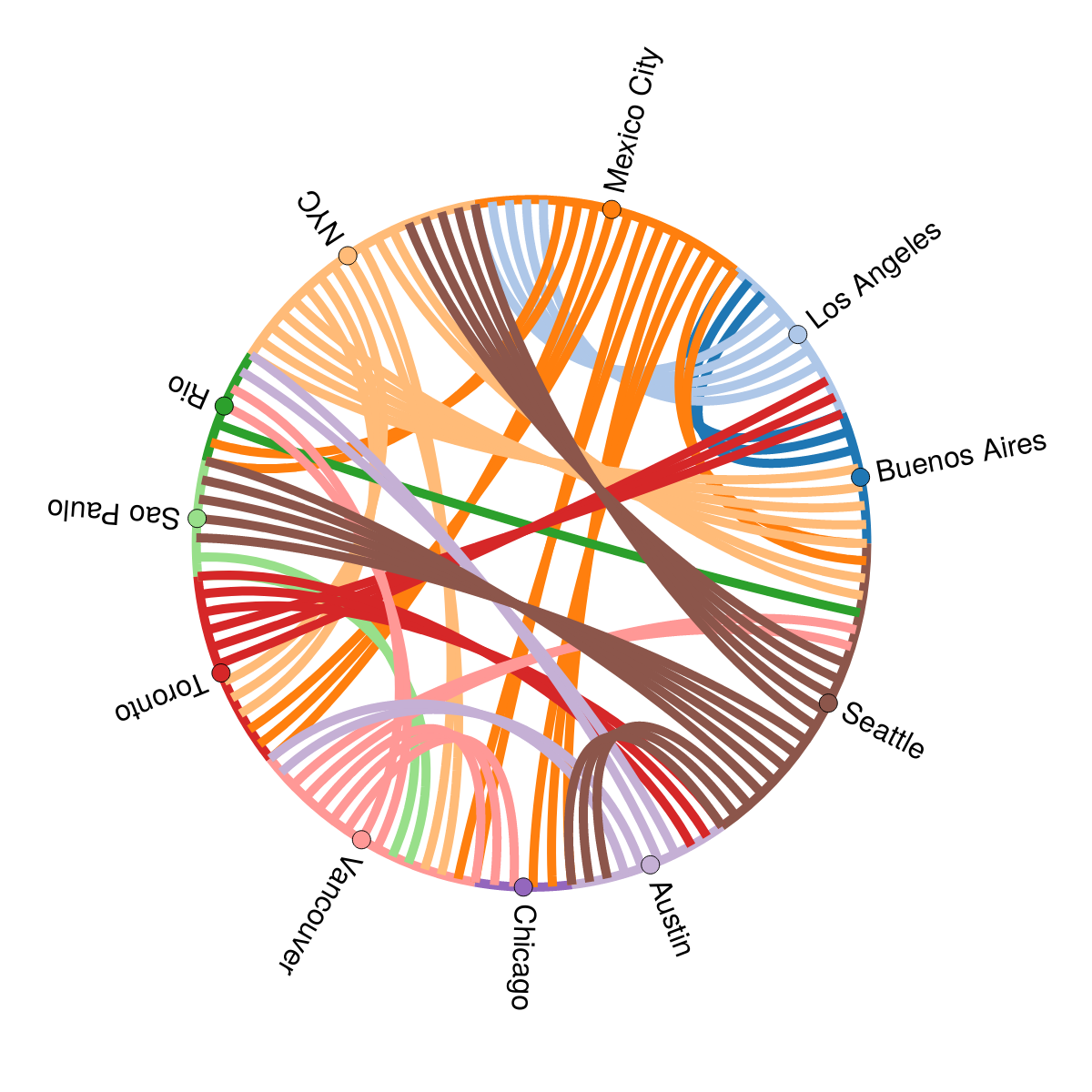}
    \label{fig:infl_cities_american}
    }\\
\subfloat[Istanbul]{
    \includegraphics[trim={1.5cm 3cm 1.5cm 1.5cm} ,clip,width=0.30\linewidth]{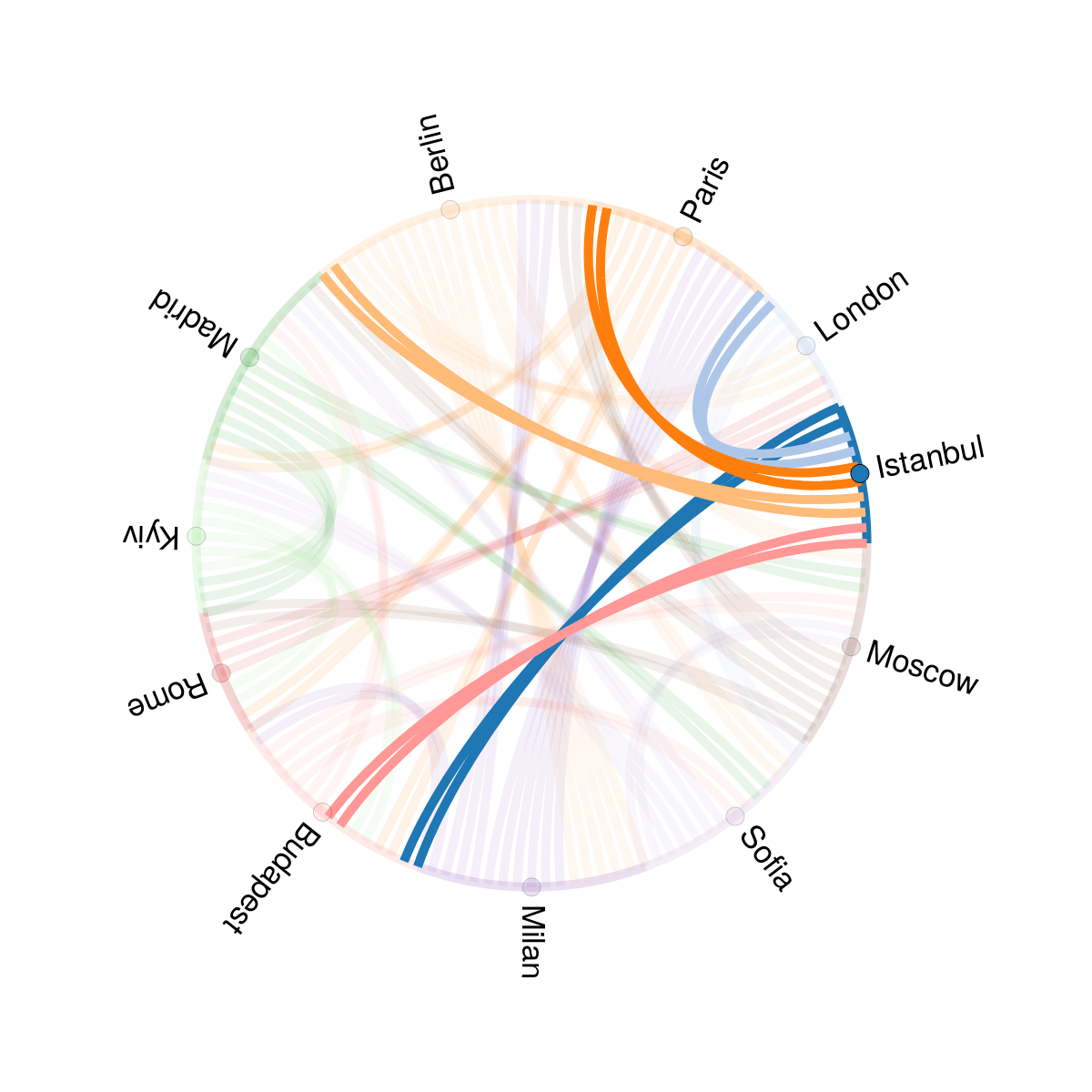}
    \label{fig:infl_cities_istanbul}
    }
\subfloat[Jakarta]{
    \includegraphics[trim={1.5cm 3cm 1.5cm 1.5cm} ,clip,width=0.30\linewidth]{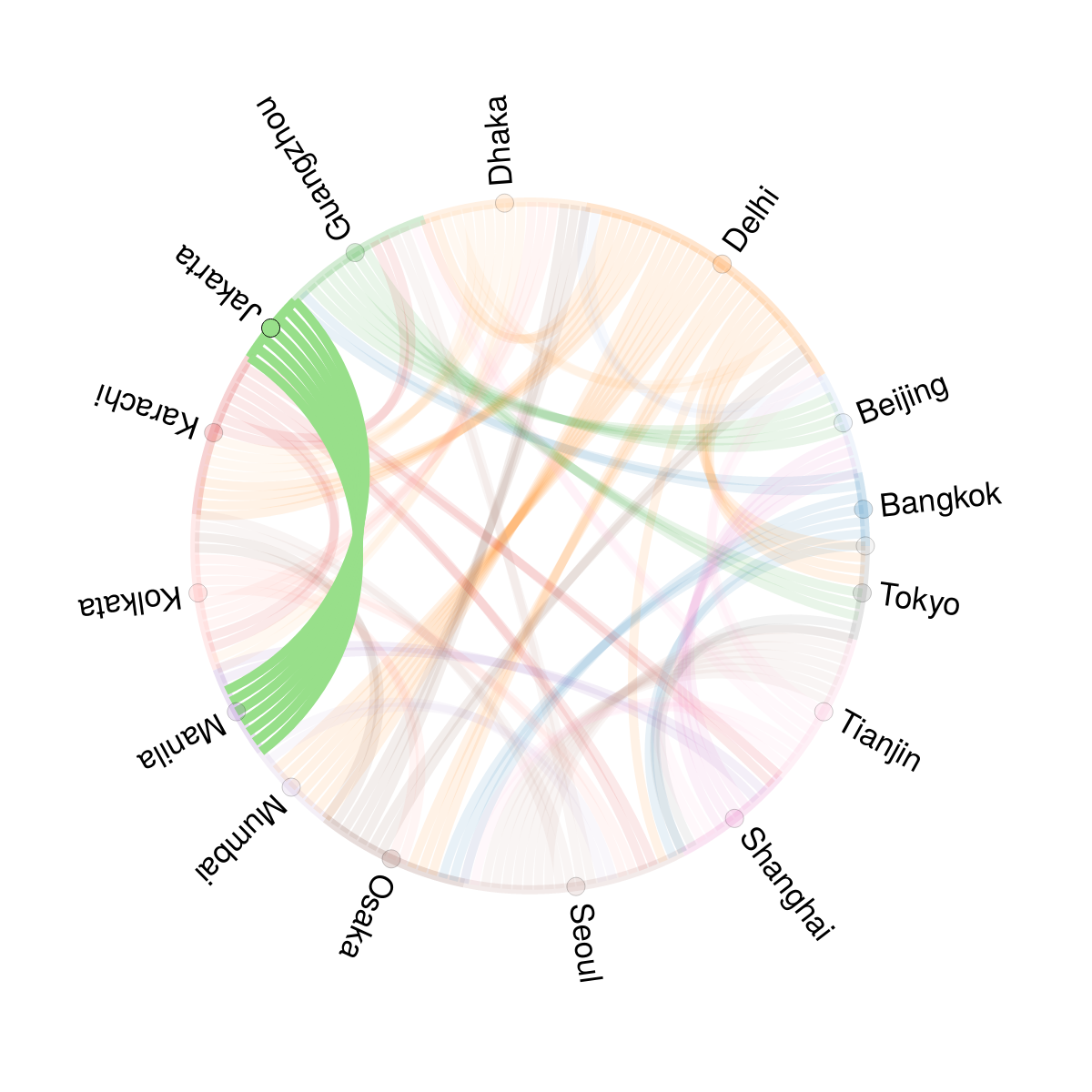}
    \label{fig:infl_cities_jakarta}
    }
\subfloat[Vancouver]{
    \includegraphics[trim={1.5cm 3cm 1.5cm 1.5cm} ,clip,width=.30\linewidth]{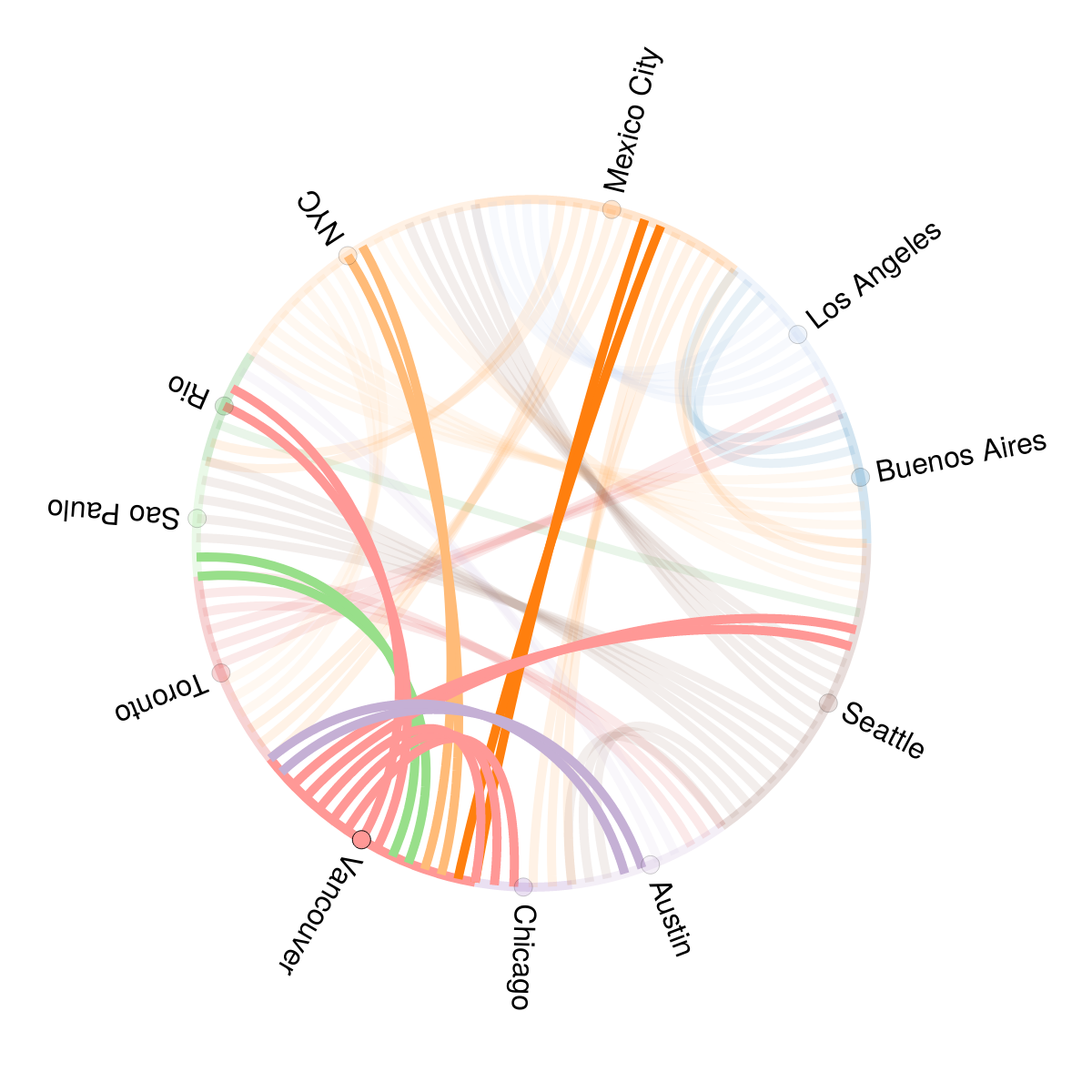}
    \label{fig:infl_cities_vancouver}
    }
\caption{
    Style influence relations discovered by our model among European (a), Asian (b) and American (c) cities.
    The number of chords coming out of a node (\ie a city) is relative to the influence weight of that city on the receiver.
    Chords are colored according to the source node color, \ie the influencer.
    Our model discovers various types of influence relations from multi-city (\eg Paris and New York City) and single-city (\eg Jakarta) influencers to cities that are mainly influence receivers (\eg Istanbul and Beijing) or influence focal points that exert and receive influence from multiple cities (\eg Vancouver).
}
\vspace{-0.3cm}
\label{fig:infl_pairwise_cities}
\end{figure*}

\begin{figure*}[!t]
\centering
\subfloat[Fashion Brands]{
    \includegraphics[width=0.33\linewidth]{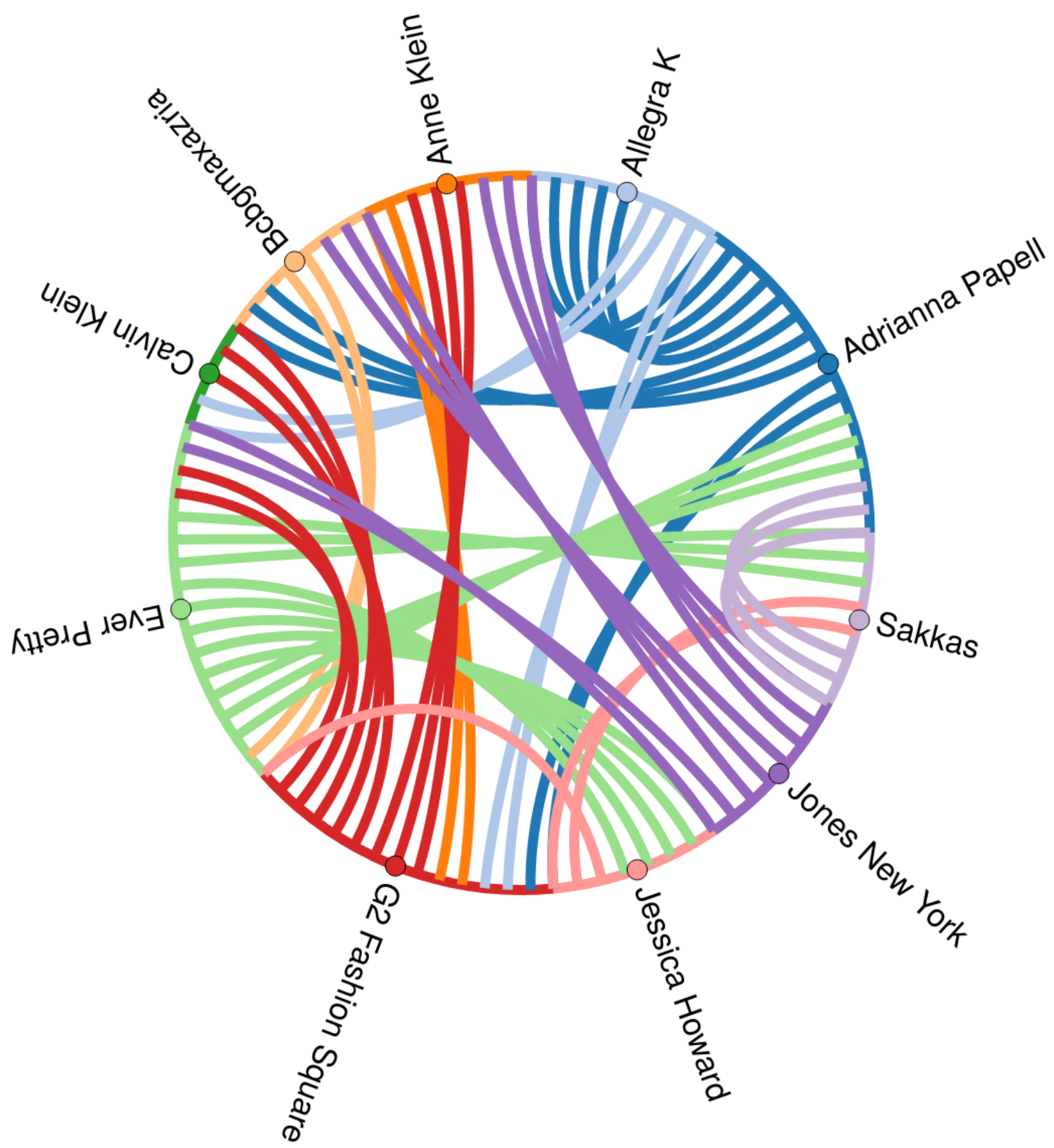}
    \label{fig:infl_brands_all}
    }
\subfloat[Calvin Klein]{
    \includegraphics[width=0.33\linewidth]{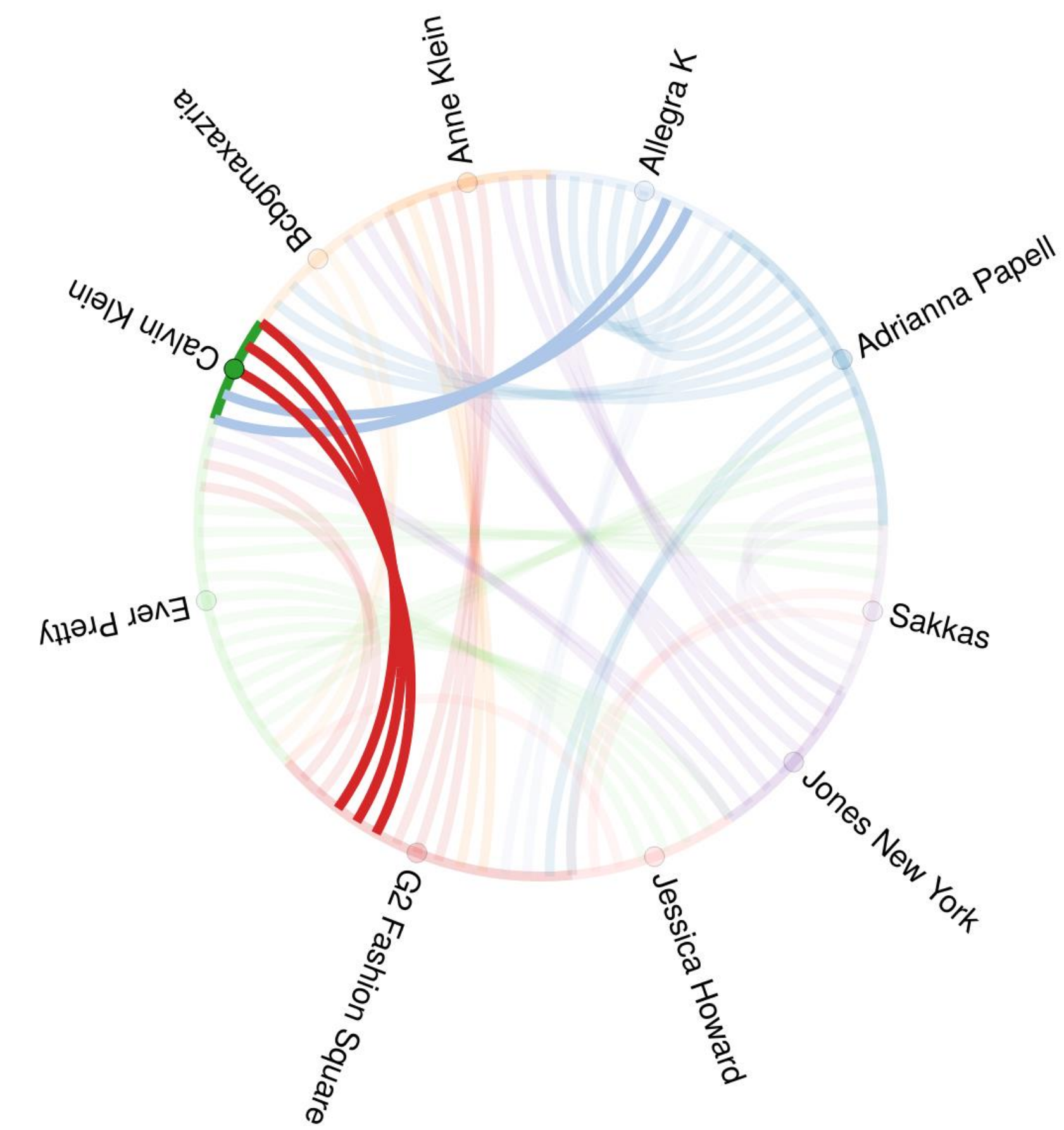}
    \label{fig:infl_brands_ck}
    }
\subfloat[Jessica Howard]{
    \includegraphics[width=0.33\linewidth]{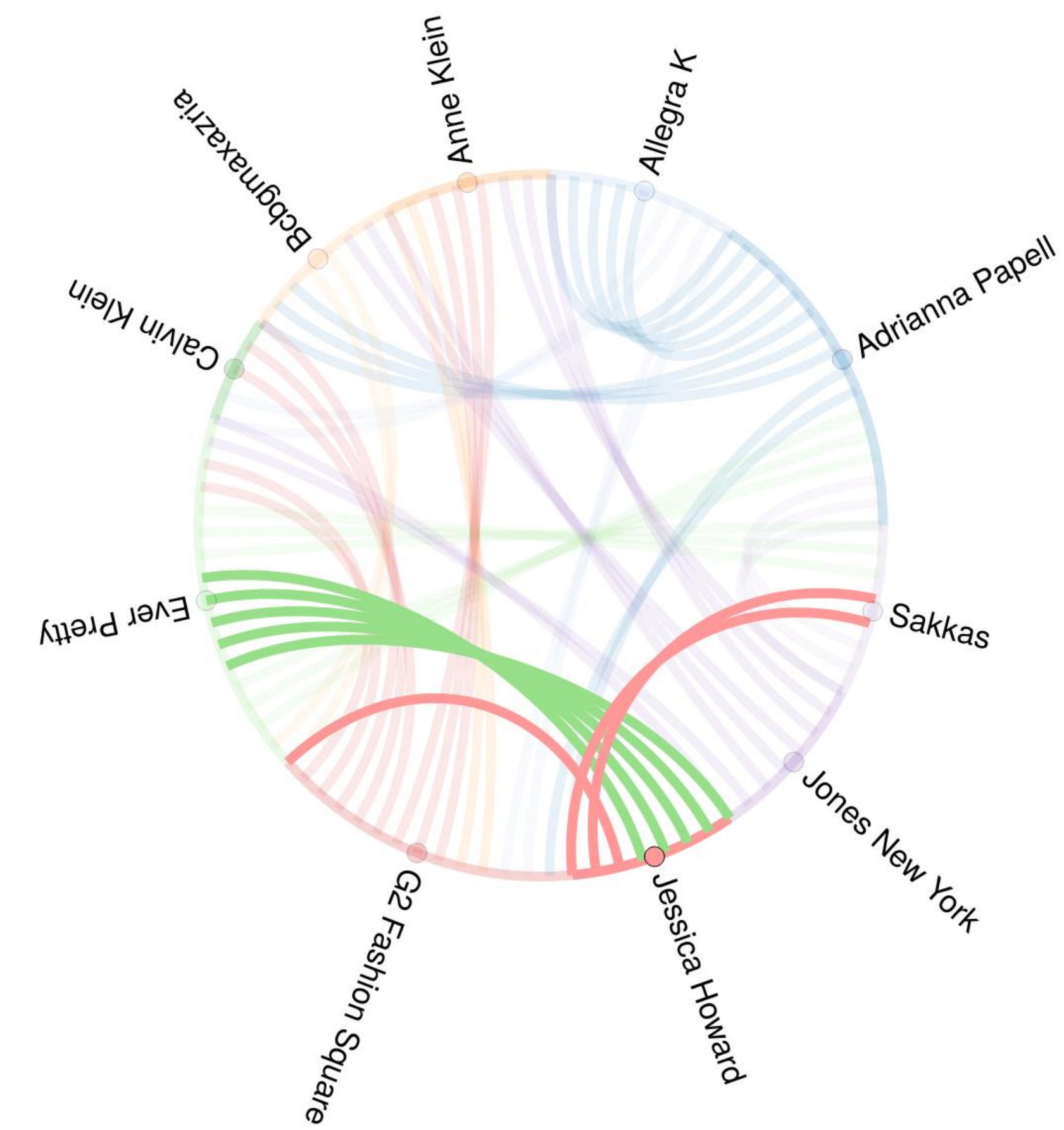}
    \label{fig:infl_brands_jessica}
    }\\
\subfloat[Jones New York]{
    \includegraphics[width=0.33\linewidth]{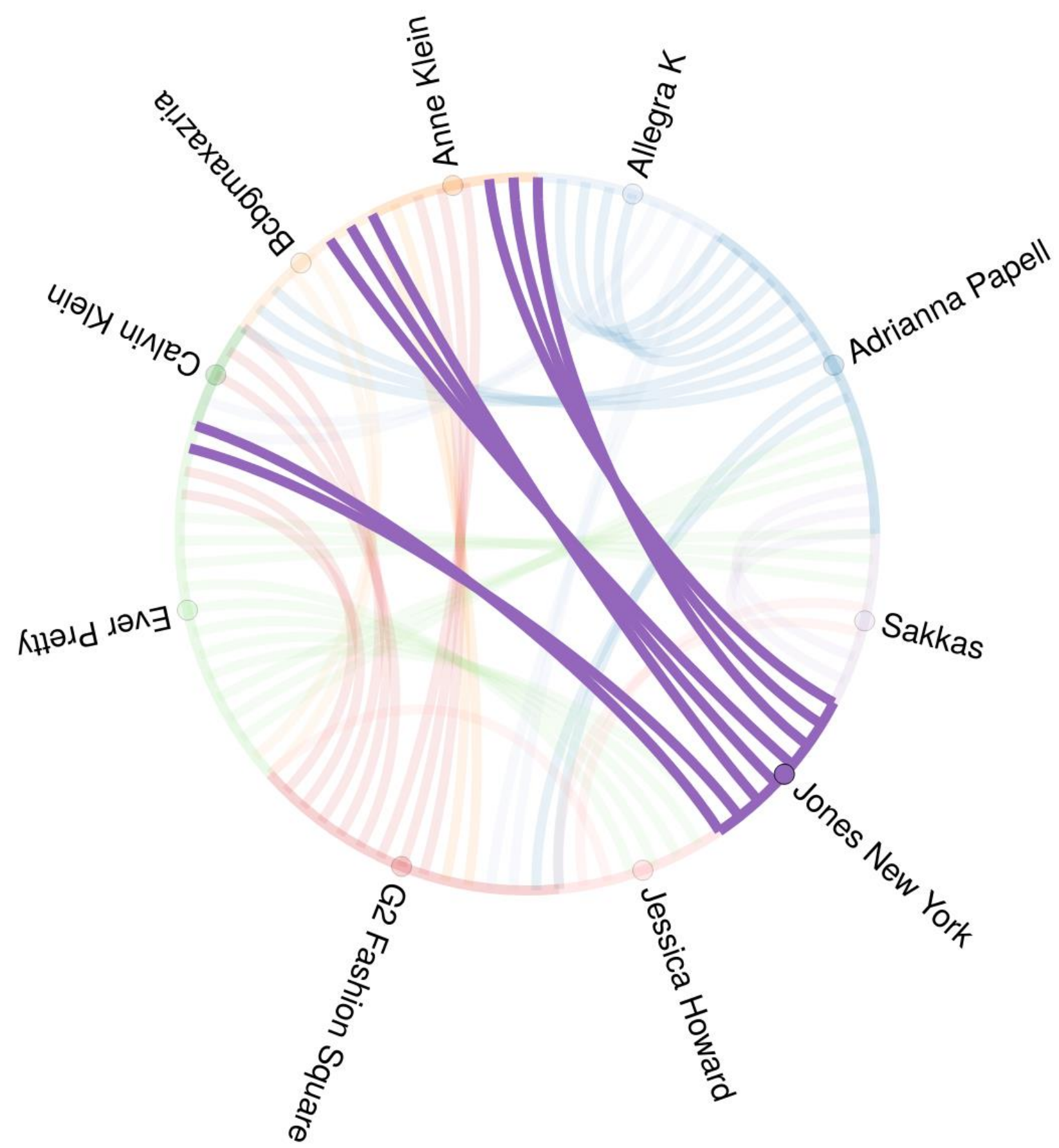}
    \label{fig:infl_brands_jonesny}
    }
\subfloat[Ever Pretty]{
    \includegraphics[width=0.33\linewidth]{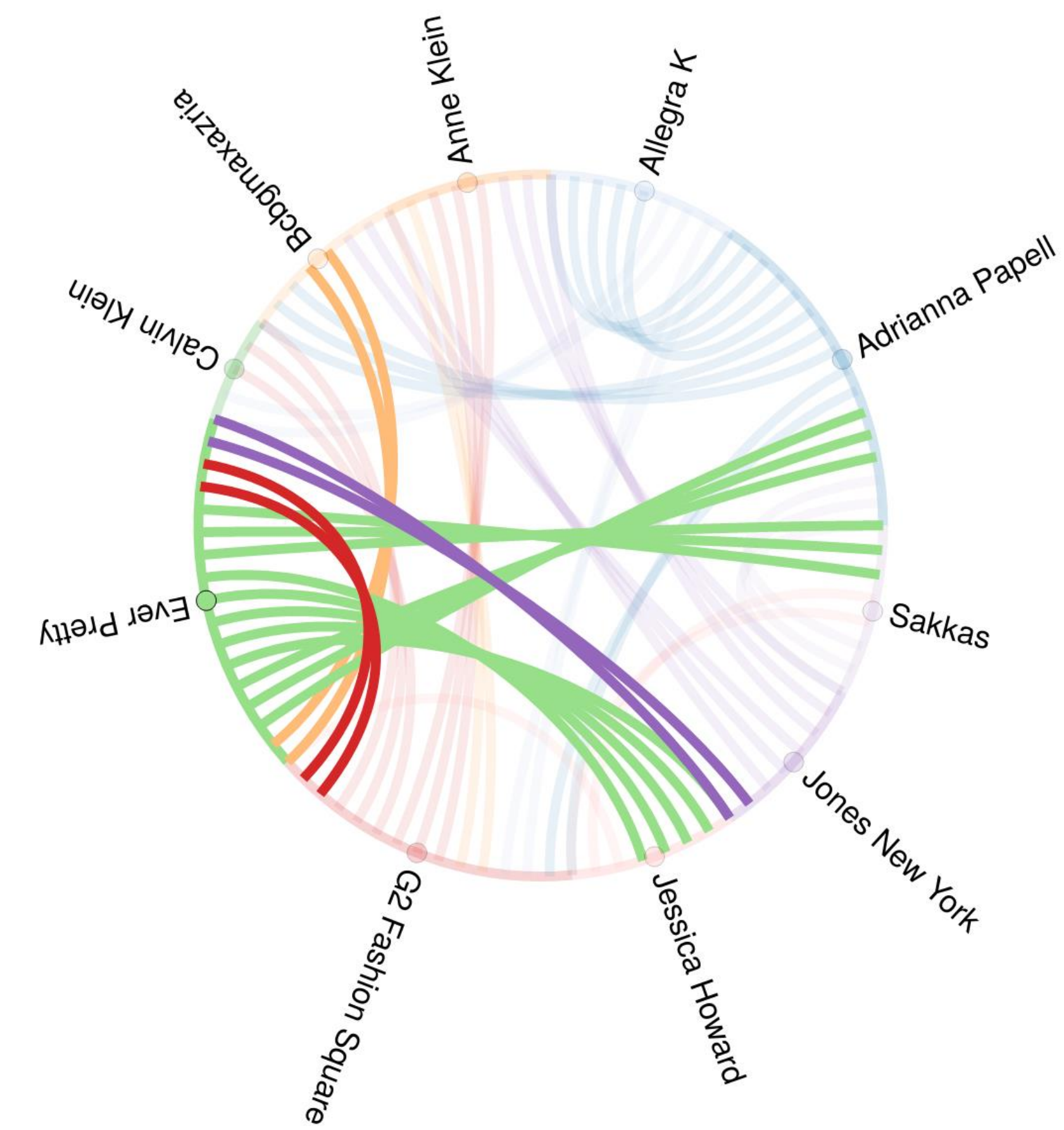}
    \label{fig:infl_brands_ever}
    }
\subfloat[Allegra K]{
    \includegraphics[width=0.33\linewidth]{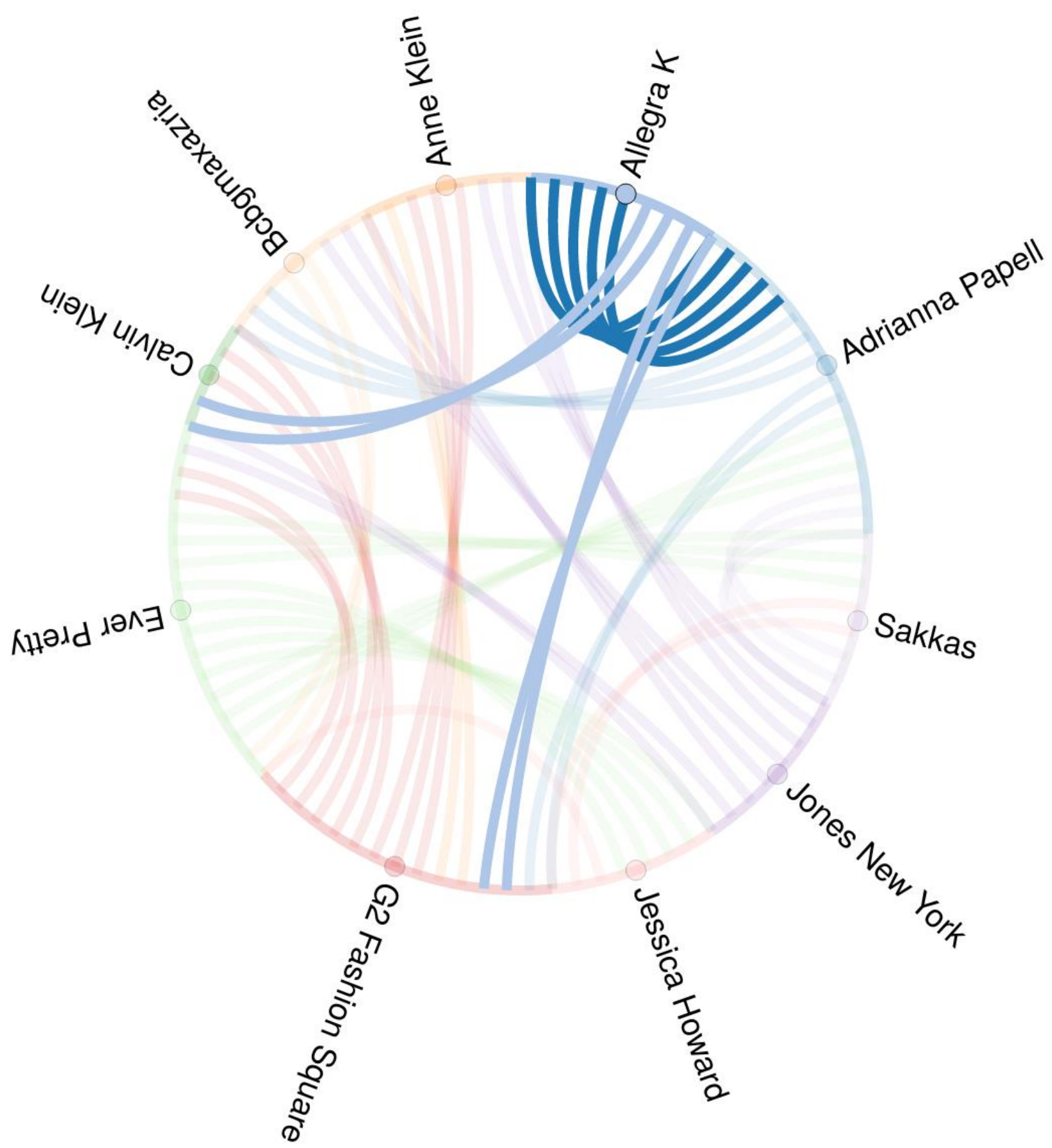}
    \label{fig:infl_brands_allegrak}
    }

\caption{
    Style influence relations discovered by our model among fashion brands (a).
    The number of chords coming out of a node (\ie a brand) is relative to the influence weight of that brand on the receiver.
    Chords are colored according to the source node color, \ie the influencer.
    While some brands have limited influence interactions (\eg Calvin Klein), others show tendency towards mainly receiving (\eg Jessica Howard) or exerting (\eg Jones New York) influence.
    Fashion brands like Ever Pretty have diverse and more balanced influence relations with the rest.
}
\vspace{-0.3cm}
\label{fig:infl_pairwise_brands}
\end{figure*}

 The ablation studies in \tblref{tbl:forecast_ablation} show the impact of each component of our model.
We evaluate our approach using only style-based (\secref{sec:app_influence_style}) or city-based (\secref{sec:app_influence_unit}) influence modeling, as well as two additional versions---one without any influence modeling, \ie it assumes a full interaction pattern among all cities, and a second version that is not trained for coherent forecasts (\secref{sec:app_forecast}).
We see that the joint modeling of the two types of influence leads to better performance.
Furthermore, we notice a larger drop in accuracy when the model does not account for influence and coherence, which shows the importance of these concepts for accurate predictions of popularity changes in the future.

Next, we analyze the learned $\alpha$ from \eqref{eq:icf} for our influence models from both datasets.
Interestingly, on GeoStyle and for the seasonal and deseasonalized splits, our model assigns equal importance for both city and style influence $\alpha=0.5$; however, on AmazonBrands the model assigns higher weight for style influence $\alpha=0.6$ compared to brand influence which is weighted with $0.4$. 
As we will see later in influence relations analysis (\secref{sec:eval_relations}) this can be attributed to the wide spread of the \emph{monopoly} type of relations between brands and styles (see \figref{fig:infl_global_brands}) which leads to a more limited direct pairwise interactions between brands compared to what we observe among cities.
This is an interesting quantitative finding that agrees with an intuition that styles ``spread" very organically between different parts of the world as people adopt new trends seen elsewhere, whereas distinct corporate brands may be more resistant to such intermingling.

\subsection{Influence Relations}
\label{sec:eval_relations}

The results thus far confirm that our method's discovered influence patterns are meaningful, as seen by their positive quantitative impact on forecasting accuracy.
Next, we analyze them qualitatively to understand more about what was learned.
We consider influence interactions along two axes: 1) a local one that looks at pairwise influence relations among the cities and the brands; and 2) a global one which examines how cities and brands influence the world's fashion trends.

\begin{figure*}[!t]
\centering
\subfloat[Asian Cities]{
    \includegraphics[width=0.47\linewidth]{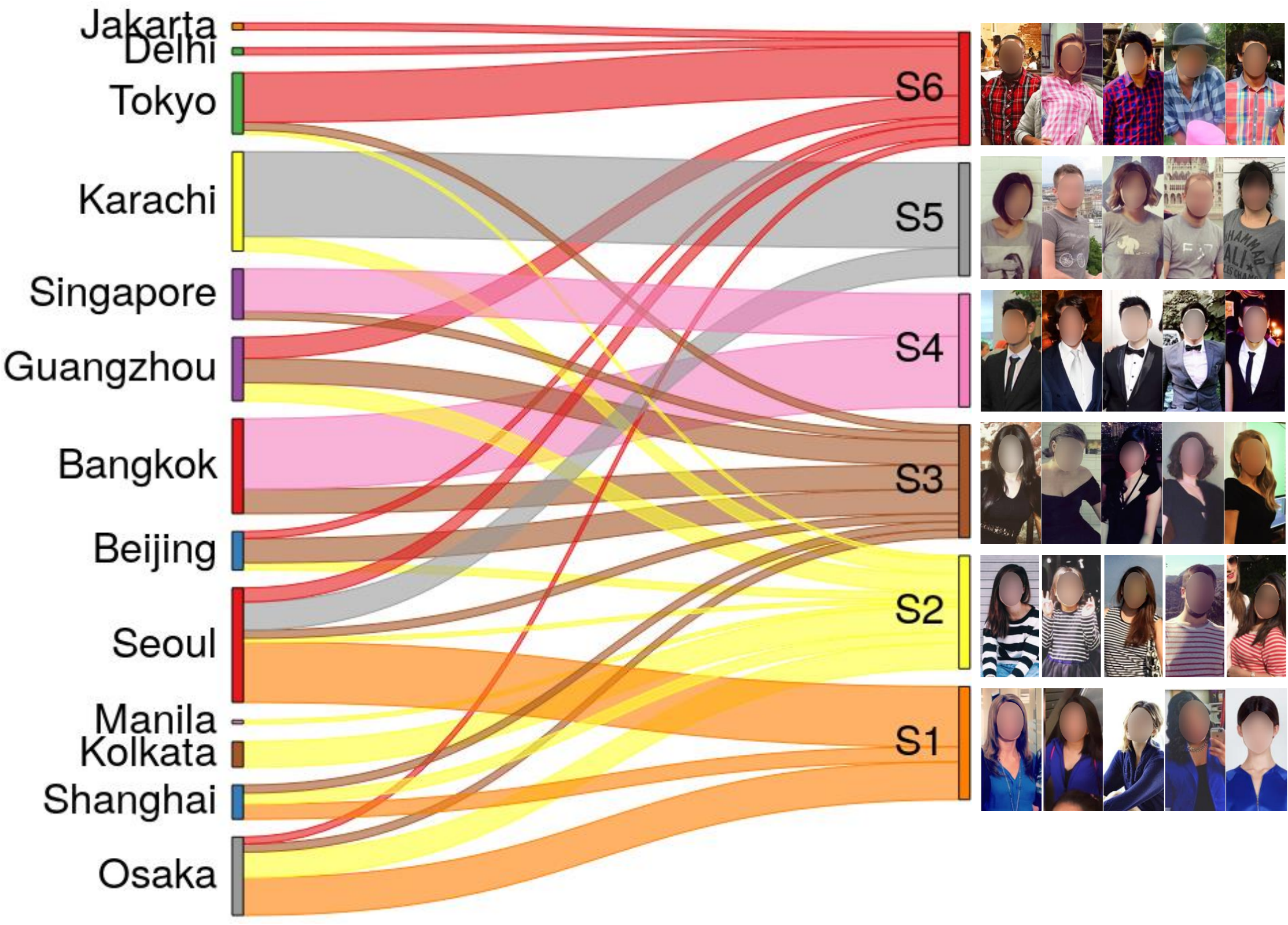}
    \label{fig:infl_global_asian}
    }\hfil
\subfloat[American Cities]{

    \raisebox{0.4cm}{\includegraphics[width=0.47\linewidth]{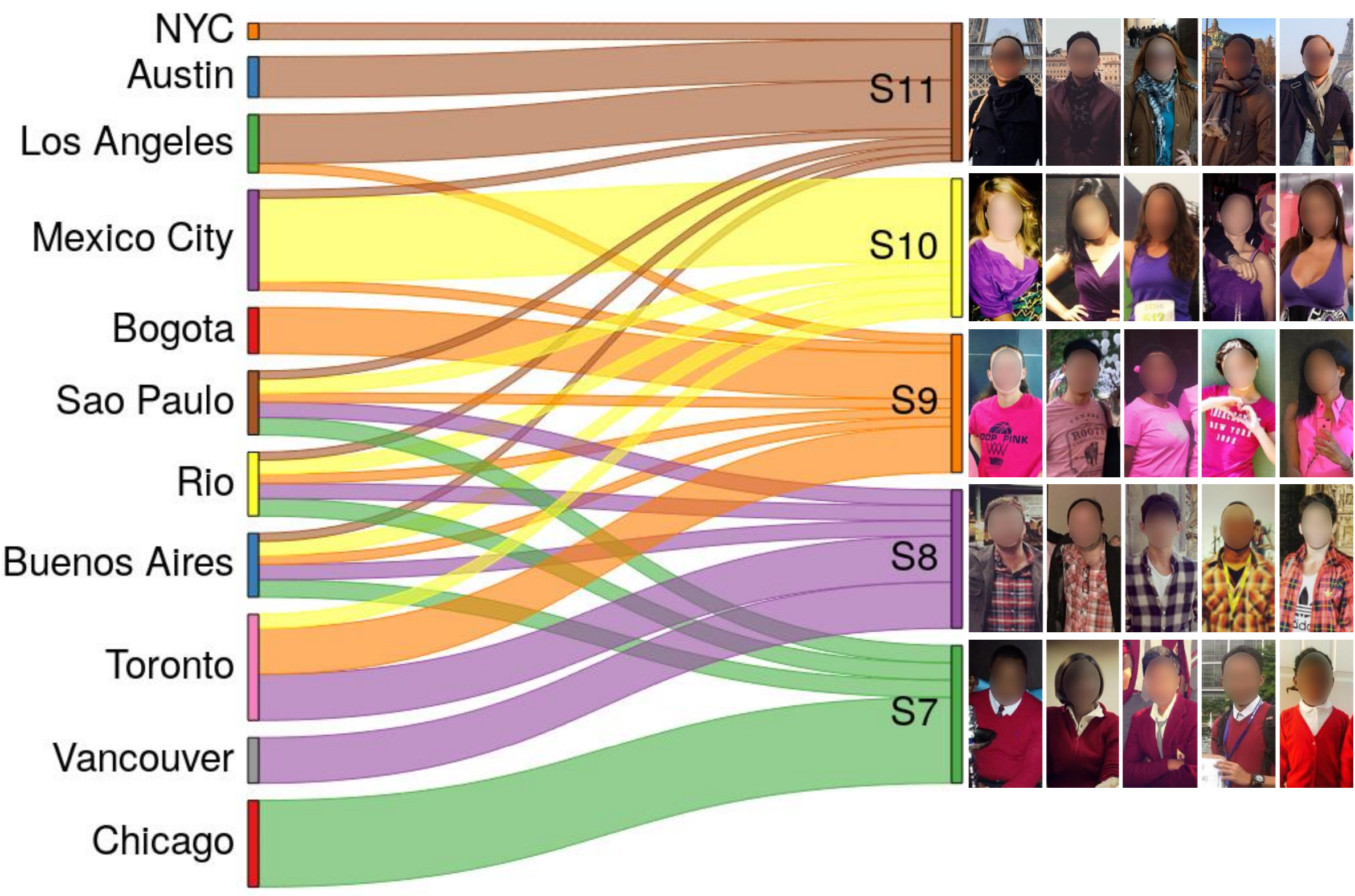}}
    \label{fig:infl_global_american}
    }
\caption{
    Discovered influence by our model of Asian (a) and American (b) cities on global trends of fashion styles learned from the GeoStyle dataset.
    The width of the connection is relative to the influence weight of that city in relation to other influencer of the same style.
}
\vspace*{-0.3cm}
\label{fig:infl_global_cities}
\end{figure*}
 
\begin{figure*}[!t]
\centering
    \includegraphics[width=.65\linewidth]{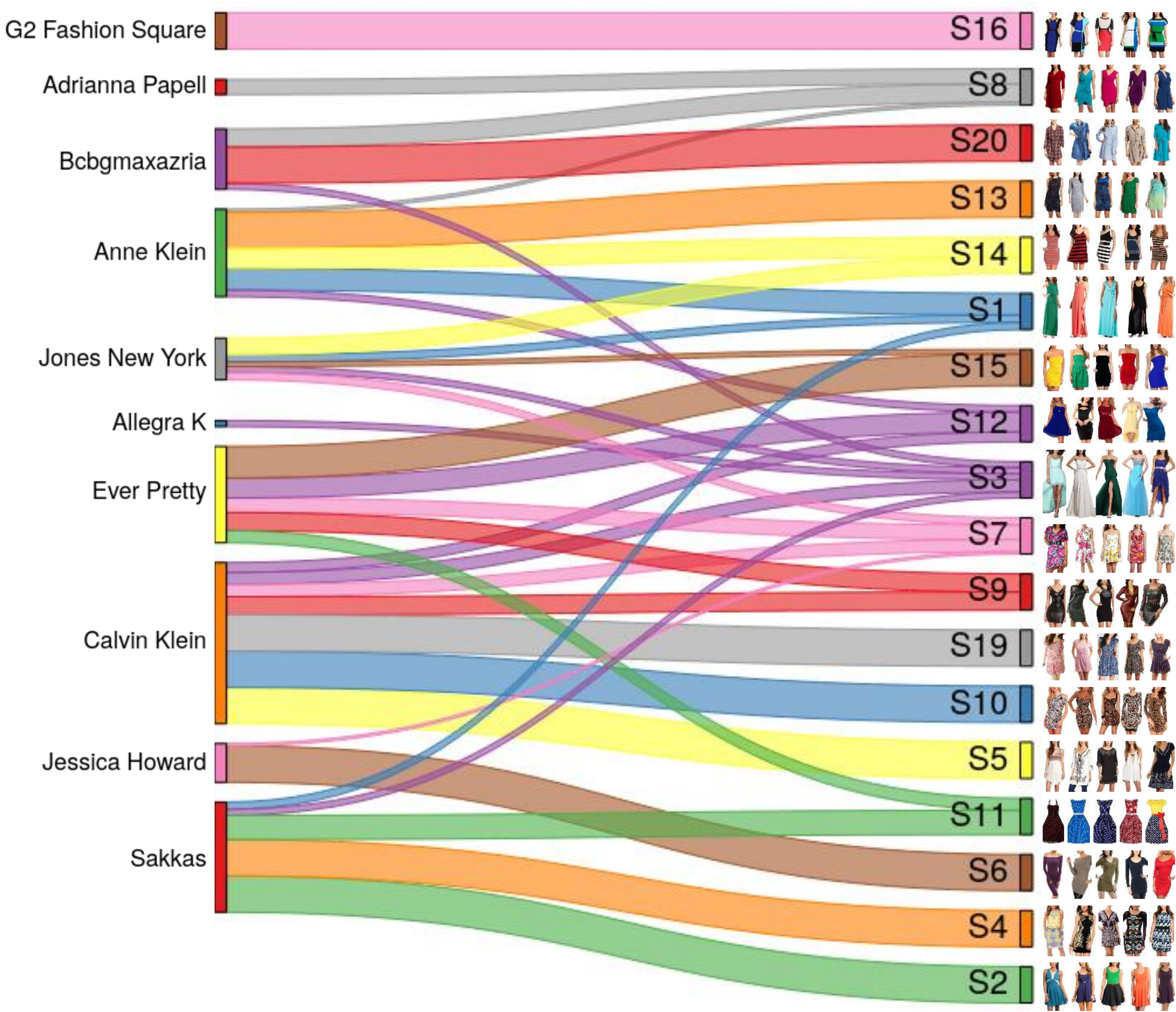}
\caption{
    Discovered influence by our model of fashion brands on global trends of $20$ dresses' fashion styles learned from the AmazonBrand dataset.
    The width of the connection is relative to the influence weight of that brand in relation to other influencers of the same style. 
}
\vspace{-0.2cm}
\label{fig:infl_global_brands}
\end{figure*}
 \parbf{1) Unit $\rightarrow$ Unit influence}
For each visual style, our model estimates the influence relation between units and at which temporal lag, yielding a tensor $\mathbf{B}\in\mathbb{R}^{|U|\times|U|\times|S|}$ such that $B^{k}_{ij}$ is the influence lag of $U^i$ on $U^j$ for style $S^k$.
By averaging these relations across all visual styles, we get an estimate of the overall influence relation between all units weighted by the temporal length, \ie long term influencers are given more weight than instantaneous ones.
We visualize the influence relations using a directed graph where each node represents a unit, and we create a weighted edge from unit $U^i$ to $U^j$ if $U^i$ is found to be influencing $U^j$.

\figref{fig:infl_pairwise_cities} shows an example of the influence pattern for fashion styles discovered by our model among major European (\figref{fig:infl_cities_european}), Asian (\figref{fig:infl_cities_asian}), and American (\figref{fig:infl_cities_american}) cities, where the number of connections between two cities is relative to the weight of the influence relation.
Our model discovers interesting patterns.
For example, there are a few fashion hubs like \emph{Paris} and \emph{Berlin} which exert influence on multiple cities while at the same time being influenced by few (one or two) cities.
\emph{Paris} influences four cities in Europe while being influenced by \emph{Milan} only (\figref{fig:infl_cities_european}).
Other cities like \emph{Vancouver} have exerted and received influence relations with multiple cities (\figref{fig:infl_cities_vancouver}).
Cities like \emph{Jakarta} have a one-to-one influence relation with \emph{Manila} (\figref{fig:infl_cities_jakarta}).
On the other end of the spectrum, we find cities like \emph{Istanbul} (\figref{fig:infl_cities_istanbul}) and \emph{Beijing} (\figref{fig:infl_cities_asian}) that mainly receive influence from multiple sources while influencing few.

\begin{figure*}[!t]
\centering
\subfloat[]{
    \includegraphics[width=.5\linewidth]{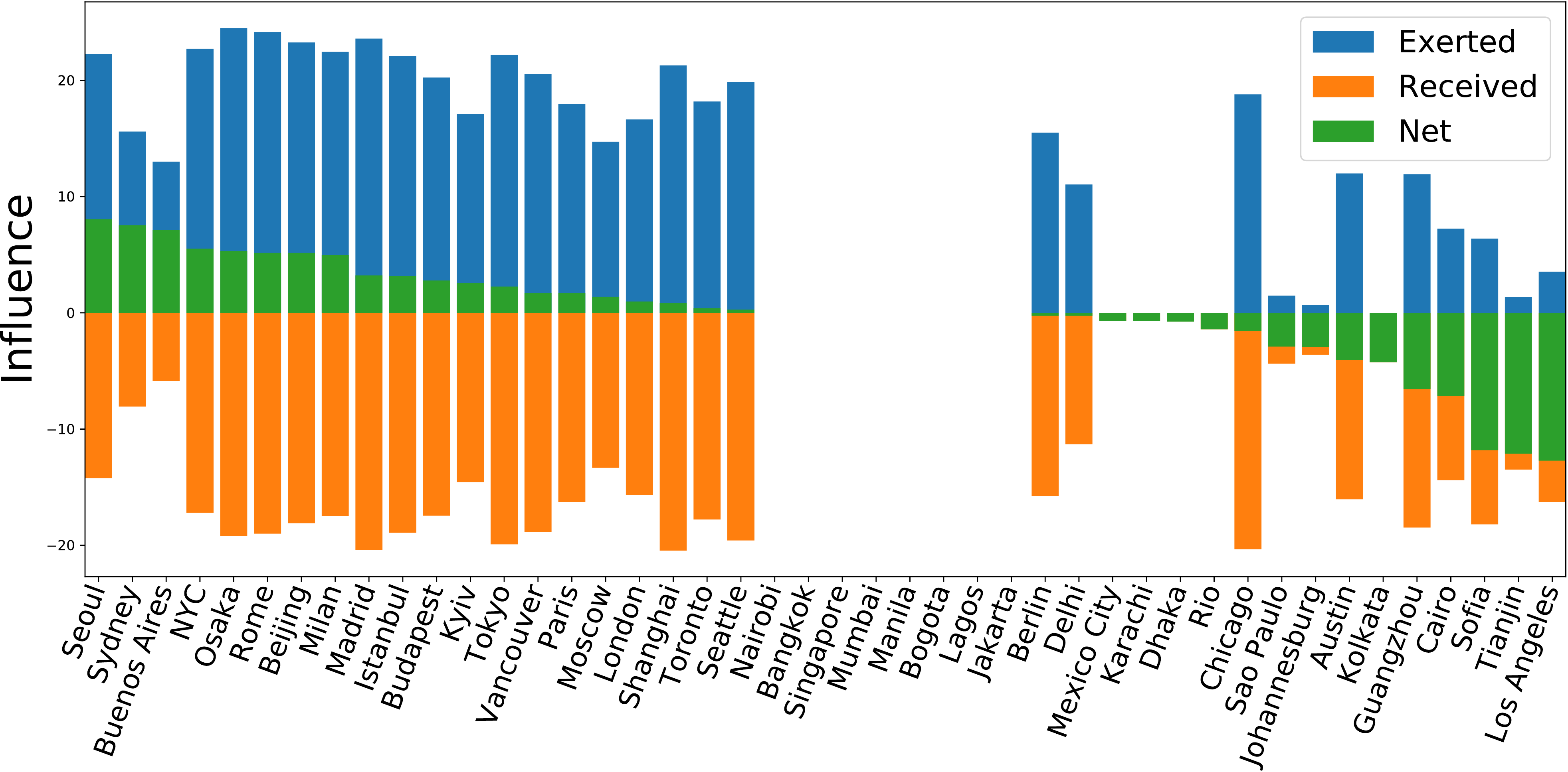}
    \label{fig:infl_rank_cities}
    }
\subfloat[]{
    \includegraphics[width=.5\linewidth]{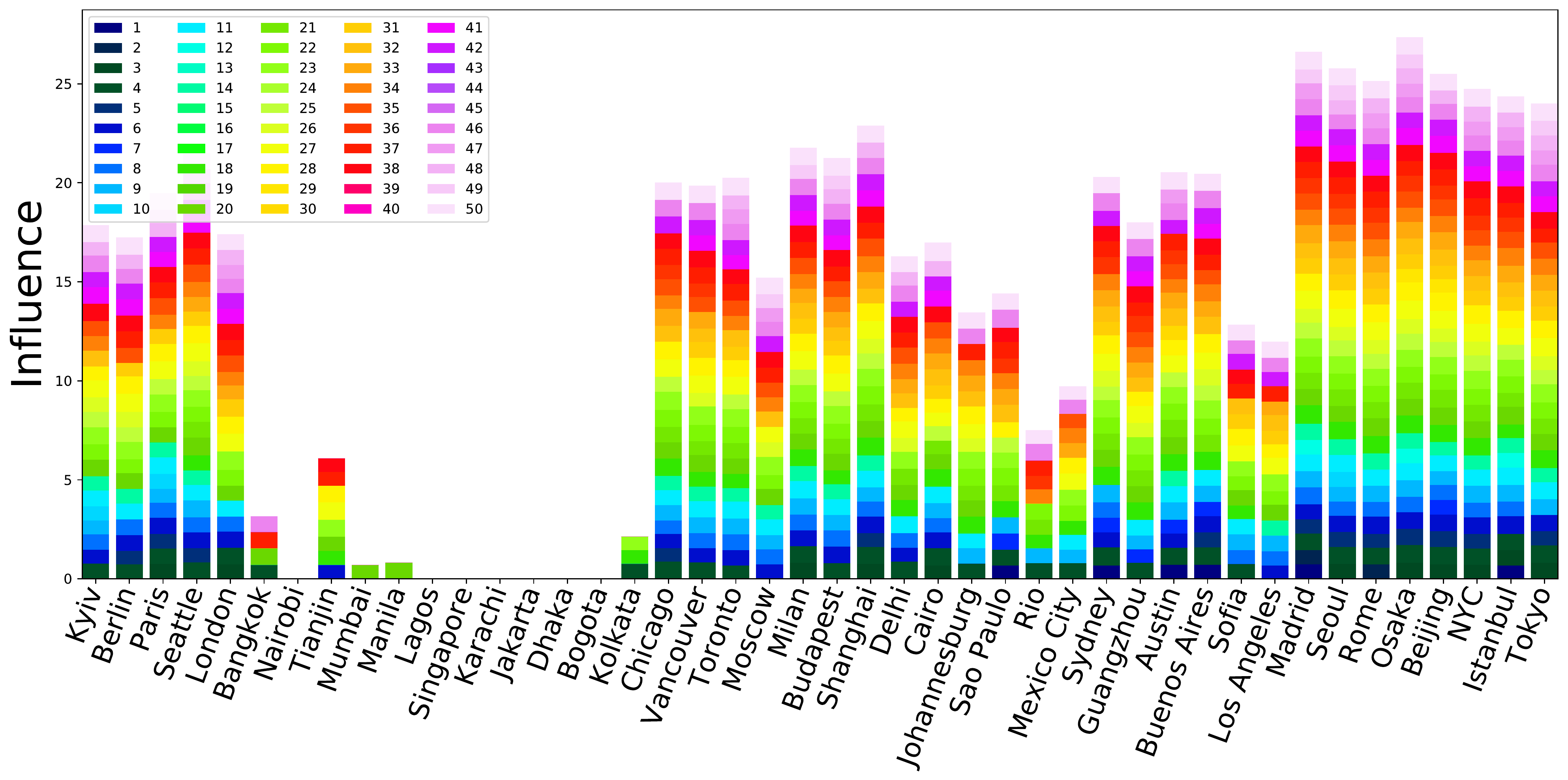}
    \label{fig:infl_rank_cities_perstyle}
    }
\\
\subfloat[]{
    \includegraphics[width=.5\linewidth]{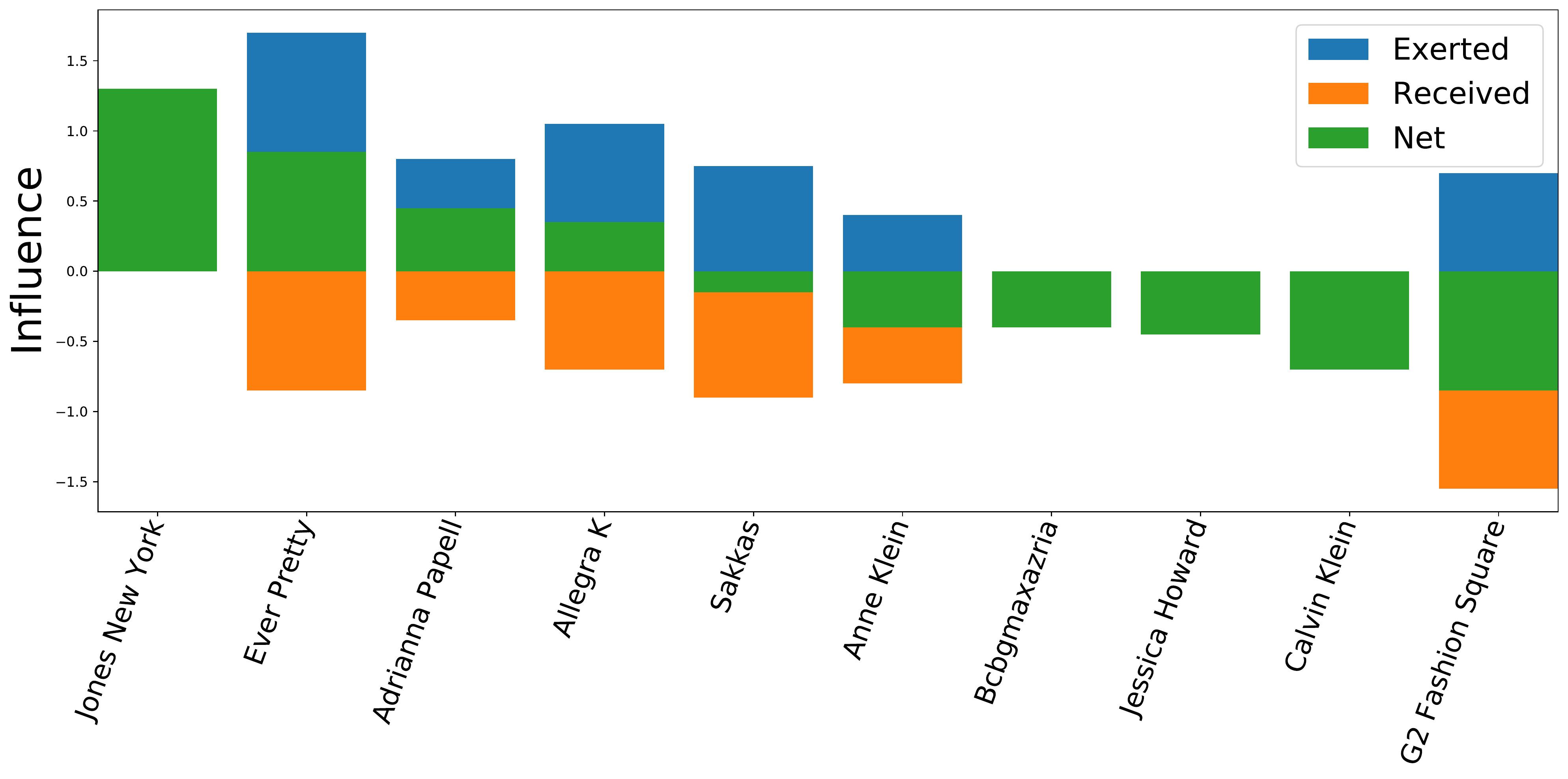}
    \label{fig:infl_rank_brands}
    }
\subfloat[]{
    \includegraphics[width=.5\linewidth]{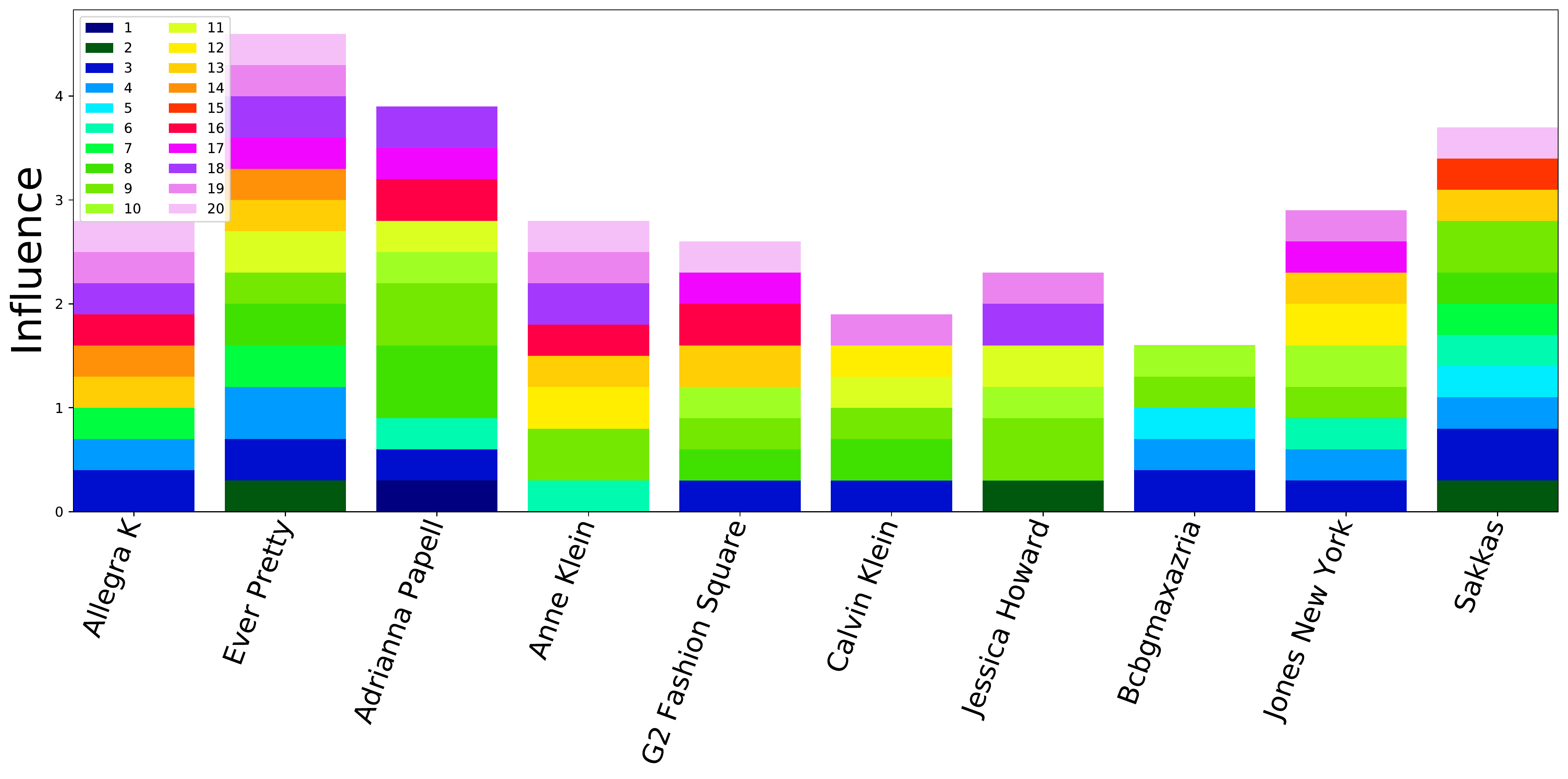}
    \label{fig:infl_rank_brands_perstyle}
    }
\caption{
    Left: Global ranking of major (a) cities and (c) brands according to their fashion influence on their peers.
    The fashion units are sorted by their \emph{net} influence score (green).
    The fashion units with no bars indicate that they do not have influential relations with above average weight.
    Right: The exerted influence score is split into influence score per individual style for each (b) city and (d) brand (sorted by style influence similarity).
}
\vspace*{-0.3cm}
\label{fig:infl_rank}
\end{figure*}
 \figref{fig:infl_pairwise_brands} shows the influence relations discovered by our model among fashion brands from the AmazonBrand dataset.
We see here similar patterns as those discovered for cities, where we can identify fashion hubs like \emph{Jones New York} (\figref{fig:infl_brands_jonesny}); influence receivers like \emph{Calvin Klein} (\figref{fig:infl_brands_ck}); and influence focal points like \emph{Ever Pretty} (\figref{fig:infl_brands_ever}).

\parbf{2) Unit $\rightarrow$ Global influence}
Alternatively, we can analyze the influence relation between a unit and the global trend for a specific style.
This helps us better understand who are the main influencers on the global stage for each of the styles.
We capture this relation by modeling the interaction of a unit's popularity trajectory on the global one (\ie the observed trend of the collective popularity of the same style across the world).

\figref{fig:infl_global_cities} shows a set of Asian (\figref{fig:infl_global_asian}) and American (\figref{fig:infl_global_american}) cities and their influence on global fashion styles.
We see that for some of the fashion styles, like $S^4$, $S^5$ and $S^7$, a couple of cities maintain a monopoly of influence on them, whereas others, like $S^3$, $S^2$ and $S^9$, are influenced almost uniformly by multiple cities.
Our influence model also reveals the influence strength (measured by the temporal lag) of these cities relative to their peers at the world stage.
See for example the strong influence of \emph{Seoul}, \emph{Bangkok}, \emph{Chicago} and \emph{Mexico City} compared to the delicate one of \emph{Manila}, \emph{Jakarta} and \emph{Sao Paulo}, as represented by the width displayed for their respective influence relation to the global trends.

\figref{fig:infl_global_brands} show the fashion brands and their influence on global fashion styles for dresses.
Interestingly, we see here that the influence monopoly pattern is more prominent than in the GeoStyle dataset.
This is expected since fashion brands usually try to develop their own signature styles, hence they are more focused on a subset of styles where they can maintain an impact on their market share.
Moreover, this view can provide us with unique information on the fashion industry.
While a brand like \emph{Calvin Klein} has little pairwise influence interaction with others (\figref{fig:infl_brands_ck}), it maintains an influence monopoly at the global scale on a larger and diverse set of styles than the rest.
We can also identify main brand competitors and on which styles.
For example, \emph{Ever Pretty} \& \emph{Calvin Klein} are the main competitors on style $S^9$, and \emph{Jones New York} and \emph{Anne Klein} on style $S^{14}$.
Other styles like $S^3$ and $S^7$ exhibit a dispersed influence pattern with no clear main influencer.

\subsection{Influence Ranking}
\label{sec:eval_ranking}

We rank all cities in the GeoStyle dataset according to their accumulated influence power on their peers.
That is, we assign an influence score for each city according to the sum of weighted influence relations \emph{exerted} by that city on the rest.
Similarly, we also calculate the sum of \emph{received} influence as well as the difference in both as the \emph{net} influence score.

\figref{fig:infl_rank_cities} shows these three influence scores for all cities across the world, sorted by the \emph{net} score.
The ranking reveals that some cities, like \emph{London} and \emph{Seattle}, act like focal points for fashion styles, \ie they receive and exert a high volume of influence simultaneously.
Others, like \emph{Seoul} and \emph{Osaka}, have a high net influence, which could indicate having some unique fashion styles not influenced by external players.
We see in \figref{fig:infl_rank_brands} a similar pattern among the fashion brands ranking, however at a lower scale compared to what is observed among the cities.
While the most influential cities may interact with up to $20$ cities, an influential brand interacts with around $1.6$ other brands on average.

\begin{figure*}[!t]
\centering
    \includegraphics[width=.85\linewidth]{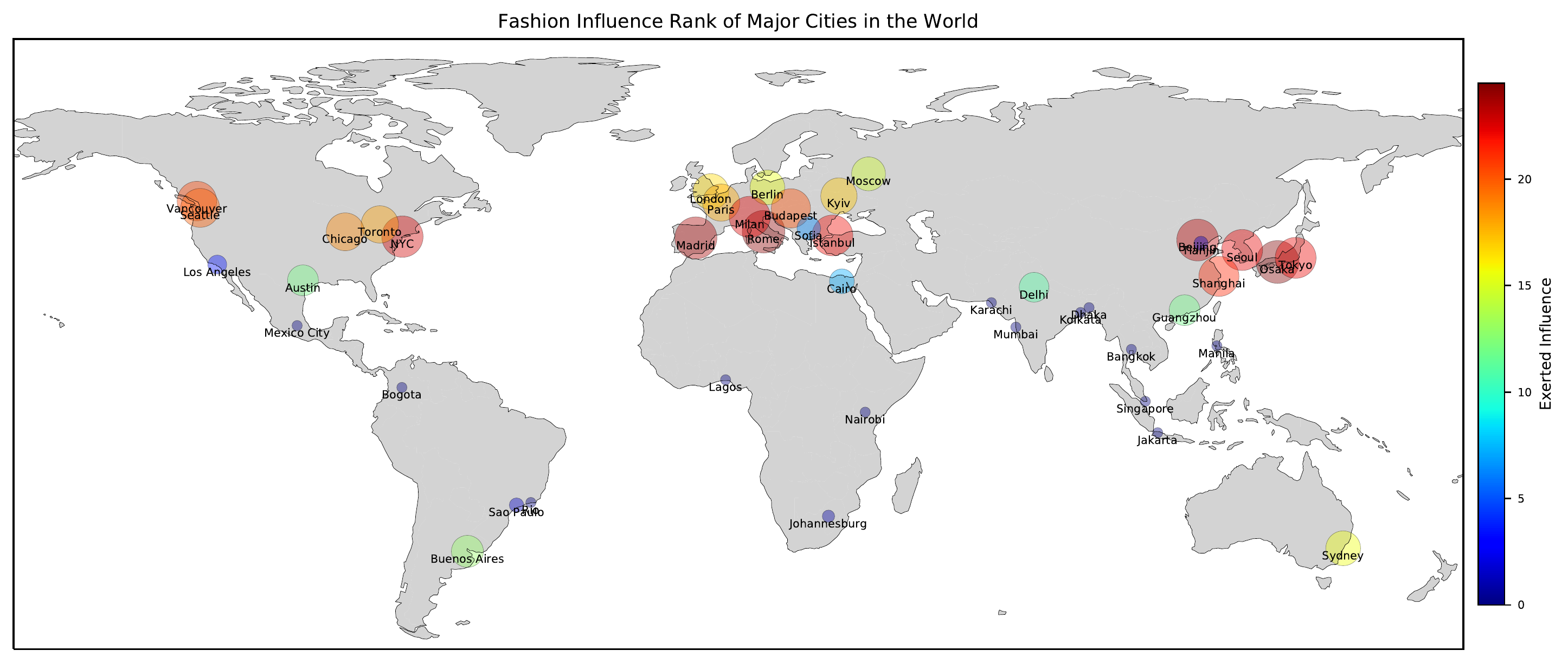}
\caption{
    Fashion influence scores as inferred by our model from everyday images of people around the world at the city level.
}
\label{fig:infl_map}
\end{figure*}
 Furthermore, breaking down the exerted influence score for each city to per-style influence scores, we see in \figref{fig:infl_rank} (right) that we can identify and group influencers into ``teams" based on their common set of styles where they exert their influence.
For example, \emph{Chicago}, \emph{Vancouver}, and \emph{Toronto} constitute a team since they seem to be influencing similar sets of visual styles (\figref{fig:infl_rank_cities_perstyle}).
Likewise, \emph{Jones New York} and \emph{Sakkas}, and \emph{Calvin Klein} and \emph{Jessica Howard} share comparable style influence profiles (\figref{fig:infl_rank_brands_perstyle}).

\figref{fig:infl_map} provides a global view of fashion influence on the world map.
It shows the exerted influence score of each city represented by its circle size and color.
We notice that most of the influential cities (among the $44$ world cities present in the GeoStyle dataset) lay in the northern hemisphere and specifically in its upper part. 
Furthermore, we aggregate the influence score inferred by our model by country.
We see that while some individual Asian and American countries have high fashion influence scores (\eg \emph{Japan}, \emph{South Korea}, and \emph{Canada}), most of the influential fashion cities and countries are located in \emph{Europe} which leads in terms of global fashion influence according to these social media images.

\subsection{Visual Influence vs.~People's Perception of Fashionability}
\label{sec:eval_perception}

Our work is the first to quantify fashion influence relations directly from large-scale visual data.
We stress that the trends visible in the photos are exactly what our model measures; there is no separate ``ground truth" against which to score the influence measurements.
However, to gain more insight into what visual fashion influence captures, we next explore how the patterns we have discovered align (or not) with existing metrics for related properties that are derived from means other than images.
In particular, we analyze the alignment between a) how our model ranks cities by their influence and b) how fashion experts rank cities by their fashionability.
``Fashionability'' means the extent to which a unit (city) exhibits the popular clothing styles.
This is a distinct property from influence, which means the extent to which a unit affects another unit's styles---here, in the Granger causality sense tied to forecastability.
Nonetheless, they are linked concepts and their degree of alignment can help us understand how well the image-driven trends our model discovers agrees with the manually-made conclusions of fashion experts and enthusiasts.\footnote{We focus this analysis on cities, as we are not aware of similar resources that succinctly record the perceived fashionability of brands.}

We consider two types of fashionability ranking:

\begin{figure*}[!t]
\centering
\subfloat[Cities]{
    \includegraphics[width=0.28\linewidth]{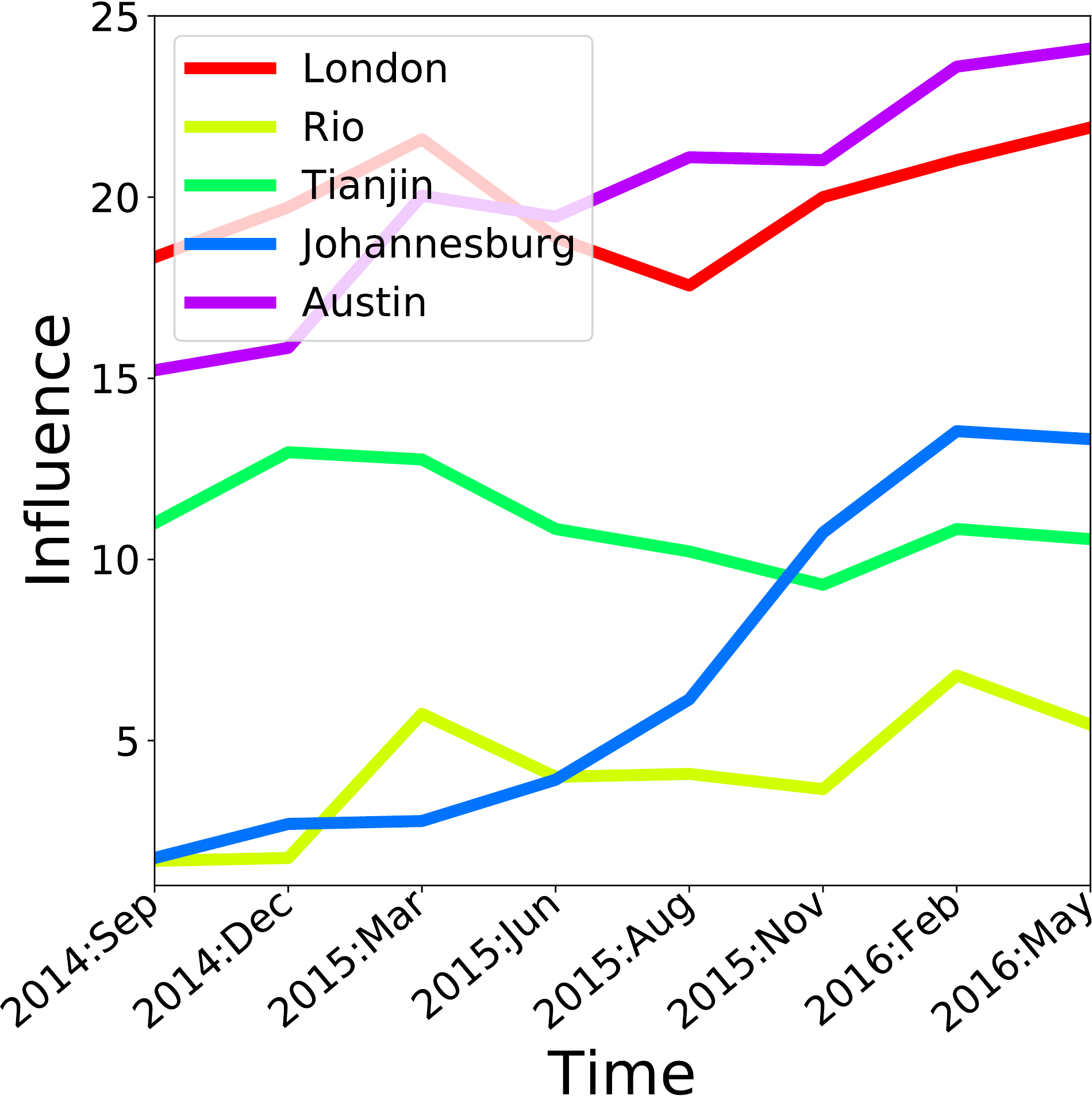} 
    \label{fig:infl_dynamics_cities}
    }\hfil
\subfloat[Fashion Brands]{
    \includegraphics[width=0.28\linewidth]{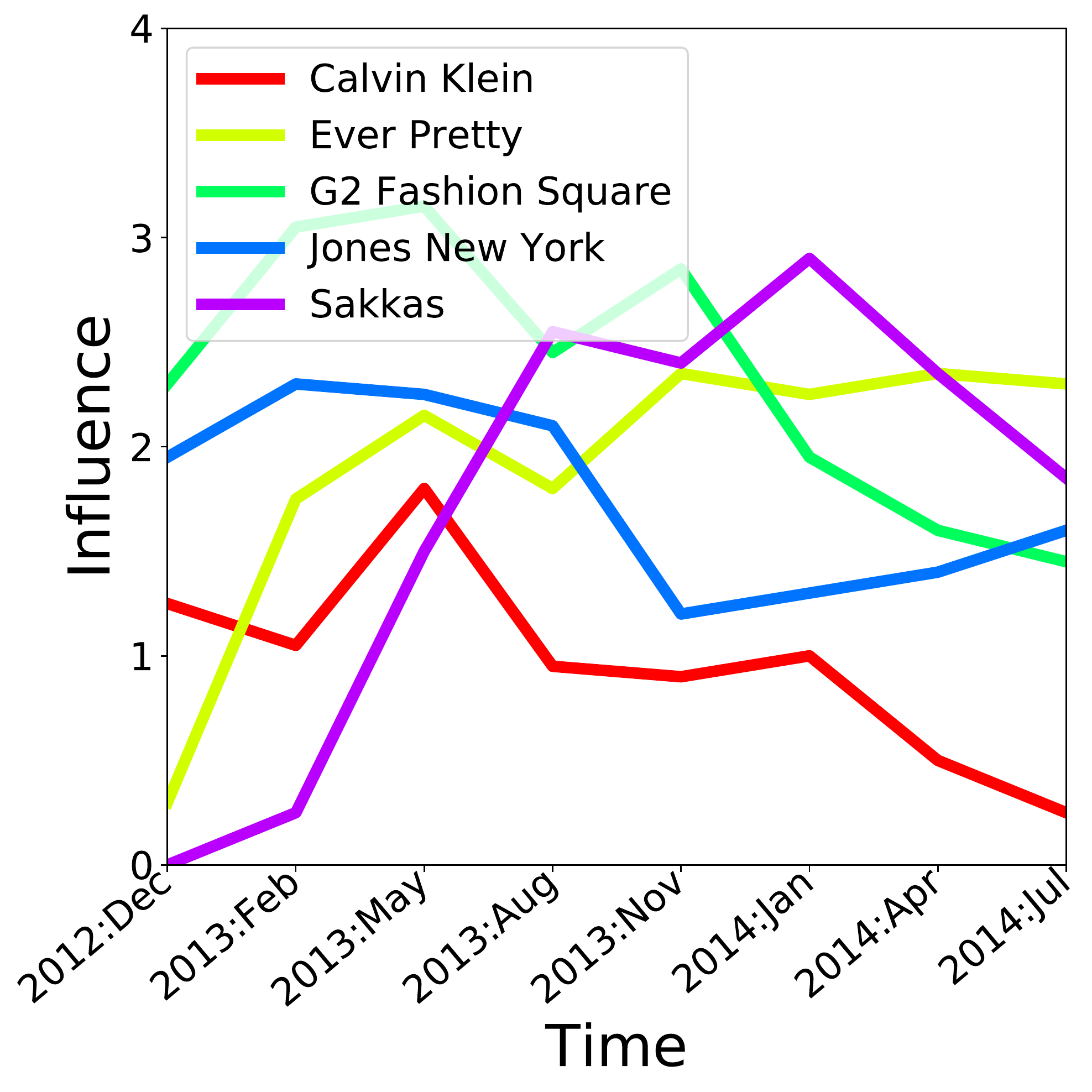}
    \label{fig:infl_dynamics_brands}
    }\hfil
\subfloat[Styles]{
    \includegraphics[width=0.39\linewidth]{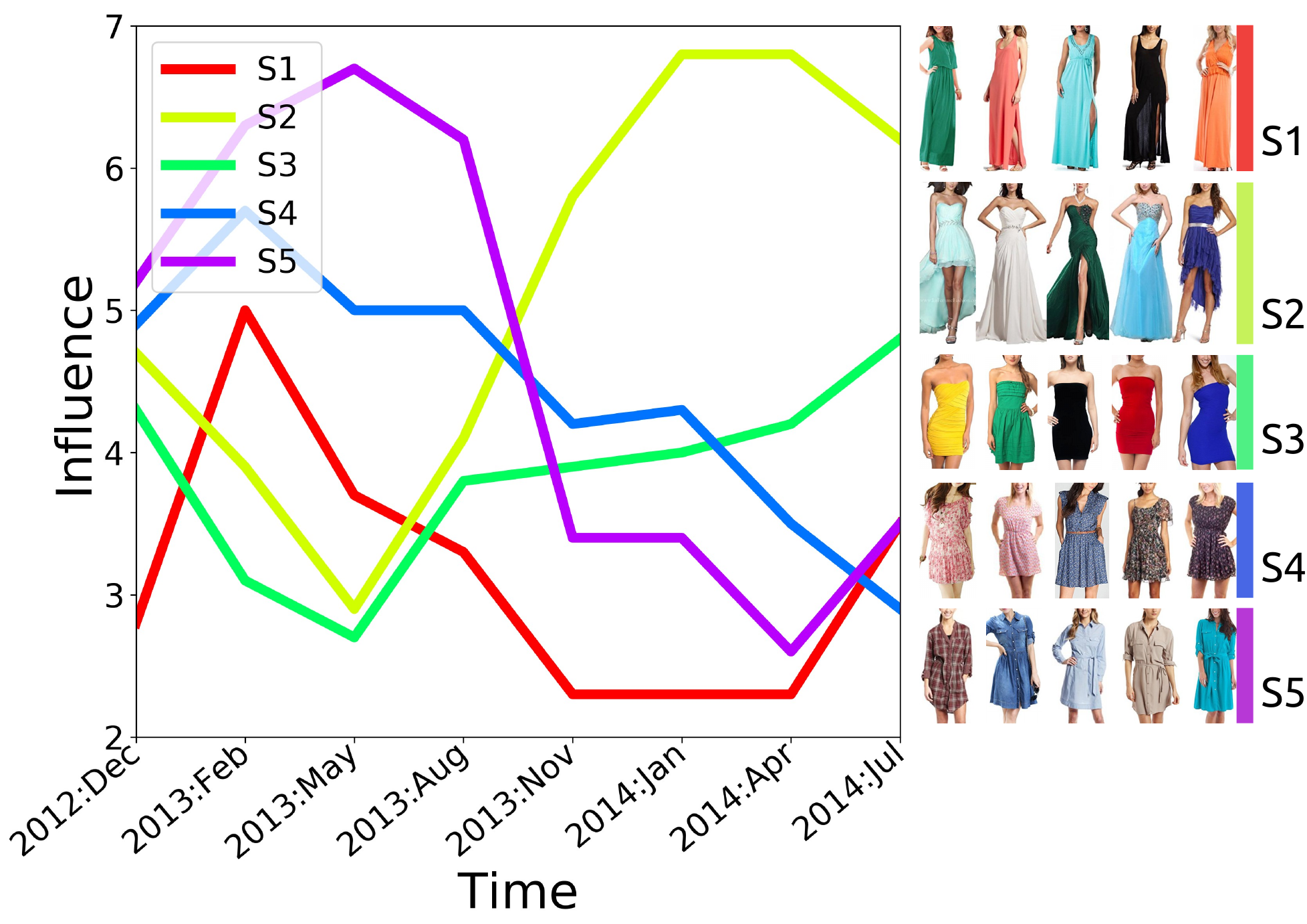}
    \label{fig:infl_dynamics_styles}
    }
\caption{
    Dynamics analysis of exerted fashion influence at multiple time steps (with a three month interval) reveals the cities' (a), the brands' (b) and the styles' (c) temporal changes in influence strength.
}
\vspace{-0.3cm}
\label{fig:infl_dynamics}
\end{figure*}
 \begin{enumerate}
\item\emph{Expert ranking}: this ranking is based on fashion journalists' assessment and takes into account factors like how many fashion shows, luxury brands head quarters, and fashion schools are located in each city.
For this type, we consider a ranking from Zalando, a fashion online retailer based in Europe, for the most elegant cities in the world~\cite{zalando_ranking}.
While this ranking also considers `urban factors' such as cleanliness and architecture, here we only consider the ranking produced based on the `fashion factors' like the ones mentioned earlier.

\item\emph{Public ranking}: this ranking is based on people's perception on which cities are more fashionable.
For this type, we consider a ranking generated by the Internet Fashion Map project~\cite{hotspots2017} that ranks each city based on how often people discuss that city in social media (\eg Twitter, Facebook) in the context of fashion using language-based statistics.
\end{enumerate}

We use the Spearman rank correlation coefficient to compare our exerted and received influence-based ranking from the GeoStyle dataset to these expert and public fashionability rankings. 
Interestingly, we find that our \emph{exerted} influence ranking is strongly aligned with public perception of fashionablity when compared to expert opinion, $0.87$ to $0.31$ respectively.
This is consistent with the data source for our model: our model discovers influence relations based on images of people's clothing from everyday activities, which makes it better aligned with public opinions in social media.
In contrast, the expert opinions are explicitly scoring cities based on high-end fashion shows and fashion from magazines.
Moreover, we find that the \emph{received} influence ranking has no correlation with fashionability, scoring only $0.05$ for both expert and public ranking.
This shows that for both the experts and the public, a fashion trend setter (\ie with high exerted influence score) is perceived as more fashionable compared to an influence receiver.

\subsection{Visual Influence Correlations with Cities' Properties}
\label{sec:eval_metacorr}

Next we analyze the correlations of influence relations discovered by our model with known real-world properties of the cities.
While the influences discovered in the photos need not align with these properties---the data-driven influences stand on their own--- it is nonetheless interesting to understand what external factors may contribute to fashion influence.
Correlating against other properties simply helps unpack what the trends do or do not relate to.

We collect information about the annual gross domestic product (GDP), the geolocation, the population size, and the yearly average temperature for each of the cities.
We calculate the correlations of these properties with the influence information discovered by our model at two levels:
1) influence world ranking (\ie does a high influence rank correlate with the population size of the city? do cities with warm weather have a high influence score?), and 2) relation direction (\ie does influence flow from high to low GDP cities? do cities influence those that are geographically close to them?).

\begin{table}[!t]
\setlength{\tabcolsep}{15pt}
\centering
\caption{
Correlations of the discovered influence patterns with meta information about the cities.
} 
\label{tbl:metacorr}
\scalebox{1.}{
\begin{tabular}{l r r}
\toprule
                    & \multicolumn{2}{c}{Fashion Influence} \\
    Meta Info.      &  Direction    &   World Rank \\
\midrule
    GDP             &   0.037       &   0.373   \\
    Temperature     &   - 0.319     &   - 0.616   \\
    Latitude        &   - 0.348     &   0.596   \\
    Population      &   0.038       &   - 0.193   \\
    Distance        &   - 0.165     &   n/a     \\
    Num. Samples    &   - 0.148     &   0.086   \\
\bottomrule
\end{tabular}
}
\vspace{-0.3cm}
\end{table}
 \parbf{World rank}
In \tblref{tbl:metacorr} (second column) we correlate the discovered influence ranking of all the cities with the ranking derived from each of the meta properties using the Spearman coefficient.
The correlation 
uncovers some curious cases.
We see a weaker but positive correlation with GDP, \ie a higher GDP could be a faint indicator of a higher fashion influence.
There is an above average correlation between the city influence rank and its latitude; many of the influential fashion cities are on the northern hemisphere.
This reinforces our previous observation from \secref{sec:eval_ranking} that most influential cities are ofter of higher latitude. 
Finally, we observe a negative correlation with average temperature; influential fashion cities are often colder.

\parbf{Relation direction}
\tblref{tbl:metacorr} (first column) shows the correlation of the influence directions discovered by our model with differences in each of the meta properties between the influencer and the influenced city.
Specifically, for each city and meta property $P_i$ (\eg GDP), first we measure the differences between that city and the rest in regards to $P_i$, then we correlate these differences with the influence exerted by that city.
Interestingly, none show high correlation with fashion influence directions.
The relation type cannot be reliably estimated based on the differences in GDP (\eg high GDP cities do not always influence lower GDP ones), population (\eg cities with high population do not necessarily influence others with lower population or vice versa), nor distance (\eg influence does not correlate well with how far one city is from its influencer).
A weak and negative correlation is found with temperature and latitude differences, showing that cities with similar temperature or at similar latitudes (\ie similar seasonal variations) tend to influence each other slightly more.
These results suggest our model discovers complex fashion influence relations that are hard to infer from generic properties of the constituent players.

As a sanity check, we also explore the correlation of the number of image samples collected from each city in the GeoStyle dataset with the two types of influence information (\tblref{tbl:metacorr} last row).
We find that there is no strong correlation between the learned influences and the number of images available in the data for each city (\ie influential cities are not those with a higher number of samples in the dataset).

\subsection{Influence Dynamics}

Finally, we study the changes in the influence rank of fashion units and styles through time.
We carry out our influence modeling based on the style trajectories of the various units and styles as before, but at multiple sequential time steps.
Then we collect the overall influence score of each unit or style at each step.

\figref{fig:infl_dynamics_cities} shows the change in the influence score for a subset of five cities spanning different continents.
We notice that cities show various dynamic behaviors across time.
While some cities like \emph{London} and \emph{Rio} maintain a steady influence score through time (at different levels), others like \emph{Austin} and \emph{Johannesburg} demonstrate a positive trend and are gaining more influence in the fashion domain over time but at varying speeds.
Other cities like \emph{Tianjin} exhibit a mild decline in their fashion influence.
Similarly, the dynamic influence analysis of fashion brands in \figref{fig:infl_dynamics_brands} reveals that some fashion brands are gaining increased influence through time like \emph{Sakkas} and \emph{Ever Pretty} while others like \emph{G2 Fashion Square} and \emph{Calvin Klein} are gradually losing some of their exerted influence.

Finally, in \figref{fig:infl_dynamics_styles} we analyze the influence changes detected by our model for a subset of visual styles from the AmazonBrands dataset.
Interestingly, we find that some of the styles like $S^1$ and $S^2$ show almost opposite trends in their influence, which may indicate a possible competition of these styles on a similar sector of customers that swing between these styles.
We also see a general decline in influence for styles like $S^4$ and $S^5$, which could indicate a broad shift in the market away from similar fashion styles.

\section{Conclusion and Future Work}

We introduced a model to quantify influence of visual fashion trends, capturing the spatio-temporal propagation of styles around the world and between major fashion brands.
Our approach integrates both fashion unit and style influence relations along with coherence regularizers to predict the future popularity of a style.
Our influence-based model outperforms state-of-the-art in forecasting methods and our analysis of the discovered influences sheds light on intriguing patterns among cities and brands.
Our findings suggest potential applications in social science and the fashion industry, where computer vision can unlock trends that are otherwise hard to capture.

We focused on this work on learning pairwise influence relations among various fashion units and styles; however, influence may exist at higher degrees of complexity.
Discovering chains of influence among multiple units may provide us with a more comprehensive view of the influence cycle and propagation, and it could help in building more compact influence-based forecasters.
Furthermore, modeling fashion influence at a finer granularity, such as individual people, could be an interesting future direction for our work.
Influencers on social media platforms 
are highly visible and interesting to users and marketers alike.
Quantifying the impact of these individual influencers on their social network and their visual style trends offers a challenging scientific problem with impact on real-world applications.

\noindent\textbf{Acknowledgements:}  We thank Utkarsh Mall for helpful input on the GeoStyle data.  UT Austin is supported in part by NSF IIS-1514118.

{\small
\bibliographystyle{ieee_fullname}
\bibliography{mybib}

\begin{thebibliography}{10}\itemsep=-1pt

\bibitem{Al-Halah2020}
Ziad Al-Halah and Kristen Grauman.
\newblock {From Paris to Berlin: Discovering Fashion Style Influences Around
  the World}.
\newblock In {\em CVPR}, 2020.

\bibitem{Al-Halah2017}
Ziad Al-Halah, Rainer Stiefelhagen, and Kristen Grauman.
\newblock {Fashion Forward: Forecasting Visual Style in Fashion}.
\newblock In {\em ICCV}, 2017.

\bibitem{banica2014neural}
Logica Banica, Daniela Pirvu, and Alina Hagiu.
\newblock Neural networks based forecasting for romanian clothing sector.
\newblock In {\em Intelligent fashion forecasting systems: Models and
  applications}, pages 161--194. Springer, 2014.

\bibitem{hotspots2017}
Thomas~Eugen Bauknecht, Jule Schaefer, and Peter Bug.
\newblock Internet fashion map: the web representation of certain fashion hot
  spots.
\newblock Technical report, Reutlingen University, Reutlingen, Germany, 2017.

\bibitem{berg2010automatic}
Tamara~L Berg, Alexander~C Berg, and Jonathan Shih.
\newblock Automatic attribute discovery and characterization from noisy web
  data.
\newblock In {\em ECCV}, 2010.

\bibitem{bossard2012apparel}
Lukas Bossard, Matthias Dantone, Christian Leistner, Christian Wengert, Till
  Quack, and Luc Van~Gool.
\newblock Apparel classification with style.
\newblock In {\em ACCV}, 2012.

\bibitem{box2015time}
George~EP Box, Gwilym~M Jenkins, Gregory~C Reinsel, and Greta~M Ljung.
\newblock {\em Time series analysis: forecasting and control}.
\newblock John Wiley \& Sons, 2015.

\bibitem{chen2018modeling}
Chengyao Chen, Zhitao Wang, Wenjie Li, and Xu Sun.
\newblock Modeling scientific influence for research trending topic prediction.
\newblock In {\em AAAI}, 2018.

\bibitem{chen2012describing}
Huizhong Chen, Andrew Gallagher, and Bernd Girod.
\newblock Describing clothing by semantic attributes.
\newblock In {\em ECCV}, 2012.

\bibitem{Chen2015}
Kuanting Chen, Kezhen Chen, Peizhong Cong, Winston~H Hsu, and Jiebo Luo.
\newblock {Who are the Devils Wearing Prada in New York City?}
\newblock In {\em ACM Multimedia}, 2015.

\bibitem{Dong_2019_ICCV}
Haoye Dong, Xiaodan Liang, Xiaohui Shen, Bochao Wang, Hanjiang Lai, Jia Zhu,
  Zhiting Hu, and Jian Yin.
\newblock Towards multi-pose guided virtual try-on network.
\newblock In {\em ICCV}, 2019.

\bibitem{granger1969}
Clive~WJ Granger.
\newblock Investigating causal relations by econometric models and
  cross-spectral methods.
\newblock {\em Econometrica: Journal of the Econometric Society}, pages
  424--438, 1969.

\bibitem{hadi2015buy}
M Hadi~Kiapour, Xufeng Han, Svetlana Lazebnik, Alexander~C Berg, and Tamara~L
  Berg.
\newblock Where to buy it: Matching street clothing photos in online shops.
\newblock In {\em ICCV}, 2015.

\bibitem{hadi2018brand}
M Hadi~Kiapour and Robinson Piramuthu.
\newblock Brand $>$ logo: Visual analysis of fashion brands.
\newblock In {\em ECCV Workshops}, 2018.

\bibitem{Han_2019_ICCV}
Xintong Han, Zuxuan Wu, Weilin Huang, Matthew~R. Scott, and Larry~S. Davis.
\newblock Finet: Compatible and diverse fashion image inpainting.
\newblock In {\em ICCV}, 2019.

\bibitem{he2016deep}
Kaiming He, Xiangyu Zhang, Shaoqing Ren, and Jian Sun.
\newblock Deep residual learning for image recognition.
\newblock In {\em CVPR}, 2016.

\bibitem{He2016}
Ruining He and Julian McAuley.
\newblock {Ups and Downs: Modeling the Visual Evolution of Fashion Trends with
  One-Class Collaborative Filtering}.
\newblock In {\em WWW}, 2016.

\bibitem{hidayati2020style}
S.~C. {Hidayati}, T.~W. {Goh}, J.~G. {Chan}, C. {Hsu}, J. {See}, W. {Lai Kuan},
  K. {Hua}, Y. {Tsao}, and W. {Cheng}.
\newblock Dress with style: Learning style from joint deep embedding of
  clothing styles and body shapes.
\newblock {\em IEEE Transactions on Multimedia}, pages 1--1, 2020.

\bibitem{hidayati2018dress}
Shintami~Chusnul Hidayati, Cheng-Chun Hsu, Yu-Ting Chang, Kai-Lung Hua,
  Jianlong Fu, and Wen-Huang Cheng.
\newblock What dress fits me best?: Fashion recommendation on the clothing
  style for personal body shape.
\newblock In {\em ACM Multimedia}, 2018.

\bibitem{hidayati2014fashion}
Shintami~C Hidayati, Kai-Lung Hua, Wen-Huang Cheng, and Shih-Wei Sun.
\newblock What are the fashion trends in new york?
\newblock In {\em ACM Multimedia}, 2014.

\bibitem{hsiao2017latent-look}
Wei-Lin Hsiao and Kristen Grauman.
\newblock Learning the latent look: Unsupervised discovery of a style-coherent
  embedding from fashion images.
\newblock In {\em ICCV}, 2017.

\bibitem{hsiao2018creating}
Wei-Lin Hsiao and Kristen Grauman.
\newblock Creating capsule wardrobes from fashion images.
\newblock In {\em CVPR}, 2018.

\bibitem{kimberly-cvpr2020}
Wei-Lin Hsiao and Kristen Grauman.
\newblock Dressing for diverse body shapes.
\newblock In {\em CVPR}, 2020.

\bibitem{hsiao2019fashion++}
Wei-Lin Hsiao, Isay Katsman, Chao-Yuan Wu, Devi Parikh, and Kristen Grauman.
\newblock Fashion++: Minimal edits for outfit improvement.
\newblock {\em ICCV}, 2019.

\bibitem{hu2015scalable}
Changwei Hu, Piyush Rai, Changyou Chen, Matthew Harding, and Lawrence Carin.
\newblock Scalable bayesian non-negative tensor factorization for massive count
  data.
\newblock In {\em Joint European Conference on Machine Learning and Knowledge
  Discovery in Databases}, pages 53--70. Springer, 2015.

\bibitem{huang2015cross}
Junshi Huang, Rogerio~S Feris, Qiang Chen, and Shuicheng Yan.
\newblock Cross-domain image retrieval with a dual attribute-aware ranking
  network.
\newblock In {\em ICCV}, 2015.

\bibitem{kataoka2019ten}
Hirokatsu Kataoka, Yutaka Satoh, Kaori Abe, Munetaka Minoguchi, and Akio
  Nakamura.
\newblock Ten-million-order human database for world-wide fashion culture
  analysis.
\newblock In {\em CVPR Workshops}, 2019.

\bibitem{kaya2014fuzzy}
Murat Kaya, Engin Ye{\c{s}}il, M~Furkan Dodurka, and Sarven S{\i}rada{\u{g}}.
\newblock Fuzzy forecast combining for apparel demand forecasting.
\newblock In {\em Intelligent fashion forecasting systems: Models and
  applications}, pages 123--146. Springer, 2014.

\bibitem{kiapour2014hipster}
M~Hadi Kiapour, Kota Yamaguchi, Alexander~C Berg, and Tamara~L Berg.
\newblock Hipster wars: Discovering elements of fashion styles.
\newblock In {\em ECCV}, 2014.

\bibitem{kim2013discovering}
Gunhee Kim and Eric Xing.
\newblock Discovering pictorial brand associations from large-scale online
  image data.
\newblock In {\em ICCV Workshops}, 2013.

\bibitem{Kuang_2019_ICCV}
Zhanghui Kuang, Yiming Gao, Guanbin Li, Ping Luo, Yimin Chen, Liang Lin, and
  Wayne Zhang.
\newblock Fashion retrieval via graph reasoning networks on a similarity
  pyramid.
\newblock In {\em ICCV}, 2019.

\bibitem{kwak2013bikers}
Iljung~S Kwak, Ana~Cristina Murillo, Peter~N Belhumeur, David~J Kriegman, and
  Serge~J Belongie.
\newblock From bikers to surfers: Visual recognition of urban tribes.
\newblock In {\em BMVC}, 2013.

\bibitem{li2017outfit}
Y. {Li}, L. {Cao}, J. {Zhu}, and J. {Luo}.
\newblock Mining fashion outfit composition using an end-to-end deep learning
  approach on set data.
\newblock {\em IEEE Transactions on Multimedia}, 19(8):1946--1955, 2017.

\bibitem{liang2016retreival}
X. {Liang}, L. {Lin}, W. {Yang}, P. {Luo}, J. {Huang}, and S. {Yan}.
\newblock Clothes co-parsing via joint image segmentation and labeling with
  application to clothing retrieval.
\newblock {\em IEEE Transactions on Multimedia}, 2016.

\bibitem{liu2012hi}
Si Liu, Jiashi Feng, Zheng Song, Tianzhu Zhang, Hanqing Lu, Changsheng Xu, and
  Shuicheng Yan.
\newblock Hi, magic closet, tell me what to wear!
\newblock In {\em ACM Multimedia}, 2012.

\bibitem{liu2012street}
Si Liu, Zheng Song, Guangcan Liu, Changsheng Xu, Hanqing Lu, and Shuicheng Yan.
\newblock Street-to-shop: Cross-scenario clothing retrieval via parts alignment
  and auxiliary set.
\newblock In {\em CVPR}, 2012.

\bibitem{liu2019styletransfer}
Y. {Liu}, W. {Chen}, L. {Liu}, and M.~S. {Lew}.
\newblock Swapgan: A multistage generative approach for person-to-person
  fashion style transfer.
\newblock {\em IEEE Transactions on Multimedia}, 21(9):2209--2222, 2019.

\bibitem{liu2016}
Ziwei Liu, Ping Luo, Shi Qiu, Xiaogang Wang, and Xiaoou Tang.
\newblock Deepfashion: Powering robust clothes recognition and retrieval with
  rich annotations.
\newblock In {\em CVPR}, 2016.

\bibitem{lu2013story}
Zheng Lu and Kristen Grauman.
\newblock Story-driven summarization for egocentric video.
\newblock In {\em CVPR}, 2013.

\bibitem{mall2019geostyle}
Utkarsh Mall, Kevin Matzen, Bharath Hariharan, Noah Snavely, and Kavita Bala.
\newblock Geostyle: Discovering fashion trends and events.
\newblock In {\em ICCV}, 2019.

\bibitem{matzen2017streetstyle}
Kevin Matzen, Kavita Bala, and Noah Snavely.
\newblock Streetstyle: Exploring world-wide clothing styles from millions of
  photos.
\newblock {\em arXiv preprint arXiv:1706.01869}, 2017.

\bibitem{mcauley2015image}
Julian McAuley, Christopher Targett, Qinfeng Shi, and Anton Van Den~Hengel.
\newblock Image-based recommendations on styles and substitutes.
\newblock In {\em ACM SIGIR Conference on Research and Development in
  Information Retrieval}, 2015.

\bibitem{misra2018decomposing}
Rishabh Misra, Mengting Wan, and Julian McAuley.
\newblock Decomposing fit semantics for product size recommendation in metric
  spaces.
\newblock In {\em ACM Conference on Recommender Systems}, 2018.

\bibitem{murillo2012urban}
Ana~C Murillo, Iljung~S Kwak, Lubomir Bourdev, David Kriegman, and Serge
  Belongie.
\newblock Urban tribes: Analyzing group photos from a social perspective.
\newblock In {\em CVPR Workshops}, 2012.

\bibitem{ren2017comparative}
Shuyun Ren, Hau-Ling Chan, and Pratibha Ram.
\newblock A comparative study on fashion demand forecasting models with
  multiple sources of uncertainty.
\newblock {\em Annals of Operations Research}, 257(1-2):335--355, 2017.

\bibitem{ren2014fashion}
Shuyun Ren, Tsan-Ming Choi, and Na Liu.
\newblock Fashion sales forecasting with a panel data-based particle-filter
  model.
\newblock {\em IEEE Transactions on Systems, Man, and Cybernetics: Systems},
  45(3):411--421, 2014.

\bibitem{shahaf2010connecting}
Dafna Shahaf and Carlos Guestrin.
\newblock Connecting the dots between news articles.
\newblock In {\em ACM SIGKDD}, 2010.

\bibitem{Simo-Serra2015}
Edgar Simo-Serra, Sanja Fidler, Francesc Moreno-Noguer, and Raquel Urtasun.
\newblock {Neuroaesthetics in Fashion: Modeling the Perception of
  Fashionability}.
\newblock In {\em CVPR}, 2015.

\bibitem{song2011predicting}
Zheng Song, Meng Wang, Xian-sheng Hua, and Shuicheng Yan.
\newblock Predicting occupation via human clothing and contexts.
\newblock In {\em ICCV}, 2011.

\bibitem{szegedy2015going}
Christian Szegedy, Wei Liu, Yangqing Jia, Pierre Sermanet, Scott Reed, Dragomir
  Anguelov, Dumitru Erhan, Vincent Vanhoucke, and Andrew Rabinovich.
\newblock Going deeper with convolutions.
\newblock In {\em CVPR}, 2015.

\bibitem{thomassey2014sales}
S{\'e}bastien Thomassey.
\newblock Sales forecasting in apparel and fashion industry: A review.
\newblock In {\em Intelligent fashion forecasting systems: Models and
  applications}, pages 9--27. Springer, 2014.

\bibitem{veit2015learning}
Andreas Veit, Balazs Kovacs, Sean Bell, Julian McAuley, Kavita Bala, and Serge
  Belongie.
\newblock Learning visual clothing style with heterogeneous dyadic
  co-occurrences.
\newblock In {\em ICCV}, 2015.

\bibitem{Vittayakorn2015}
Sirion Vittayakorn, Kota Yamaguchi, Alexander~C. Berg, and Tamara~L. Berg.
\newblock {Runway to realway: Visual analysis of fashion}.
\newblock In {\em WACV}, 2015.

\bibitem{wang2018toward}
Bochao Wang, Huabin Zheng, Xiaodan Liang, Yimin Chen, Liang Lin, and Meng Yang.
\newblock Toward characteristic-preserving image-based virtual try-on network.
\newblock In {\em ECCV}, 2018.

\bibitem{Wang_2018_CVPR}
Wenguan Wang, Yuanlu Xu, Jianbing Shen, and Song-Chun Zhu.
\newblock Attentive fashion grammar network for fashion landmark detection and
  clothing category classification.
\newblock In {\em CVPR}, 2018.

\bibitem{Yu_2019_ICCV}
Cong Yu, Yang Hu, Yan Chen, and Bing Zeng.
\newblock Personalized fashion design.
\newblock In {\em ICCV}, 2019.

\bibitem{Yu_2019_CVPR}
Weijiang Yu, Xiaodan Liang, Ke Gong, Chenhan Jiang, Nong Xiao, and Liang Lin.
\newblock Layout-graph reasoning for fashion landmark detection.
\newblock In {\em CVPR}, 2019.

\bibitem{zalando_ranking}
Zalando.
\newblock The most elegant cities in the world.
\newblock \url{https://www.zalando.co.uk/worlds-most-elegant-cities/}.
\newblock Accessed April 7, 2020.

\bibitem{zhang2017recommendation}
X. {Zhang}, J. {Jia}, K. {Gao}, Y. {Zhang}, D. {Zhang}, J. {Li}, and Q. {Tian}.
\newblock Trip outfits advisor: Location-oriented clothing recommendation.
\newblock {\em IEEE Transactions on Multimedia}, 19(11):2533--2544, 2017.

\bibitem{Zhao_2017_CVPR}
Bo Zhao, Jiashi Feng, Xiao Wu, and Shuicheng Yan.
\newblock Memory-augmented attribute manipulation networks for interactive
  fashion search.
\newblock In {\em CVPR}, 2017.

\end{thebibliography}
}

\end{document}